\documentclass{article} % For LaTeX2e
\usepackage{iclr2027_conference,times}

\usepackage{amsmath,amsfonts,bm}

\def\eqref#1{equation~\ref{#1}}
\def\1{\bm{1}}

\def\vc{{\bm{c}}}

\def\vv{{\bm{v}}}

\def\vx{{\bm{x}}}

\def\vz{{\bm{z}}}

\DeclareMathAlphabet{\mathsfit}{\encodingdefault}{\sfdefault}{m}{sl}
\SetMathAlphabet{\mathsfit}{bold}{\encodingdefault}{\sfdefault}{bx}{n}

\usepackage[utf8]{inputenc}
\usepackage[T1]{fontenc}
\usepackage[table]{xcolor}
\usepackage{amsmath,amsfonts,amssymb,amsthm}
\usepackage{booktabs,multirow,colortbl,array}
\usepackage{graphicx,wrapfig,subcaption}
\usepackage{algorithm,algpseudocode}
\usepackage{nicefrac,microtype,pifont,xspace,soul}
\usepackage{url}
\usepackage{hyperref}
\usepackage{dsfont}
\usepackage{wrapfig}
\usepackage{wrapfig}
\usepackage{fix-cm}
\usepackage{marvosym}   % provides \Letter (envelope icon)

\algrenewcommand\algorithmicrequire{\textbf{Input:}}
\algrenewcommand\algorithmicensure{\textbf{Output:}}

\definecolor{darkgreen}{RGB}{0,128,0}
\definecolor{darkred}{RGB}{175,0,0}
\definecolor{lightblue}{HTML}{87CEFA}
\definecolor{lightpurple}{HTML}{8470FF}
\definecolor{myblue}{RGB}{38,145,199}
\definecolor{mygreen}{RGB}{38,199,149}
\definecolor{Gray}{gray}{0.9}
\definecolor{Better}{rgb}{0.18,0.407,0.266}
\definecolor{Worse}{rgb}{0.35,0.35,0.35}
\definecolor{darkpurple}{rgb}{0.38,0.27,0.61}
\definecolor{granate}{rgb}{0.64,0.16,0.16}
\definecolor{myorange}{HTML}{FFA500}

\newcommand{\ours}{PACT\xspace}
\renewcommand{\vv}{\mathbf{v}}
\renewcommand{\vz}{\mathbf{z}}

\renewcommand{\vx}{\mathbf{x}}

\renewcommand{\vc}[2]{$#1_{\text{\tiny$\pm#2$}}$}
\newcommand{\Cknown}{\mathcal{C}_{\mathrm{known}}}
\newcommand{\Cnovel}{\mathcal{C}_{\mathrm{novel}}}
\newcommand{\colKnown}{Known}
\newcommand{\colNovel}{Novel}
\newcommand{\nrm}[1]{\frac{#1}{\lVert #1\rVert_2}}
\newcommand{\mypar}[1]{\noindent\textbf{#1}\hspace{0.10mm}}
\newcolumntype{C}[1]{>{\centering\arraybackslash}p{#1}}

\newcommand{\cmark}{\textcolor{green!60!black}{\ding{51}}}
\newcommand{\xmark}{\textcolor{red}{\ding{55}}}

\title{Test-Time Generalized Category Discovery}

\author{%
\textbf{Shambhavi Mishra}\textsuperscript{1,\Letter} \quad
\textbf{Omprakash Chakraborty}\textsuperscript{1} \quad
\textbf{Julio Silva-Rodr\'{i}guez}\textsuperscript{1} \\
\textbf{Ismail Ben Ayed}\textsuperscript{1} \quad
\textbf{Marco Pedersoli}\textsuperscript{1} \quad
\textbf{Jose Dolz}\textsuperscript{1} \\[4pt]
\textsuperscript{1}\'{E}TS Montr\'{e}al, Canada
}
\iclrfinalcopy % Uncomment for camera-ready version, but NOT for submission.
\begin{document}

\maketitle
\lhead{Preprint. Under review.}   % if you used this option
{\renewcommand{\thefootnote}{}%
 \footnotetext{\Letter\ Corresponding author: \texttt{shambhavi.mishra.1@ens.etsmtl.ca}}}
 
\begin{abstract}
Test-Time Adaptation (TTA) and Generalized Category Discovery (GCD) are traditionally treated as disjoint problems: the former adapts models to domain shift assuming all test classes are known, while the latter discovers novel categories assuming labeled training data for known classes. However, real-world deployment rarely fits either setting. Motivated by this gap, we introduce \textbf{Test-Time Generalized Category Discovery (TT-GCD)}, a unified and more realistic scenario where a vision-language model must adapt to distribution shifts, classify known categories using only textual supervision, and discover novel categories, all during test time and without access to labeled data. To address this challenging scenario, we propose \textbf{\ours}(\textbf{P}rototype \textbf{A}ssignment for \textbf{C}ategory discovery at \textbf{T}est time), a fully unsupervised framework that casts known-class recognition and novel-class discovery via prototype assignment. \ours\ first re-aligns shifted visual features with the text-derived class representations of the VLM using confident zero-shot predictions. Known and novel categories are then both represented by prototypes in the visual embedding space, estimated from the unlabeled test stream, and each test image is assigned to the category whose prototype is most similar to its visual feature.
%Known and novel categories are then represented uniformly by visual prototypes estimated from the unlabeled test stream, and classified by a single nearest-prototype rule. 
Extensive experiments across corruption and domain-shift benchmarks demonstrate that \ours\ outperforms adapted state-of-the-art TTA and GCD methods, effectively bridging the gap between adaptation and discovery.
\end{abstract}

\section{Introduction}
\begin{figure}[!h]
    % \vspace{-20pt}
    \centering
    \includegraphics[width=0.9\linewidth]{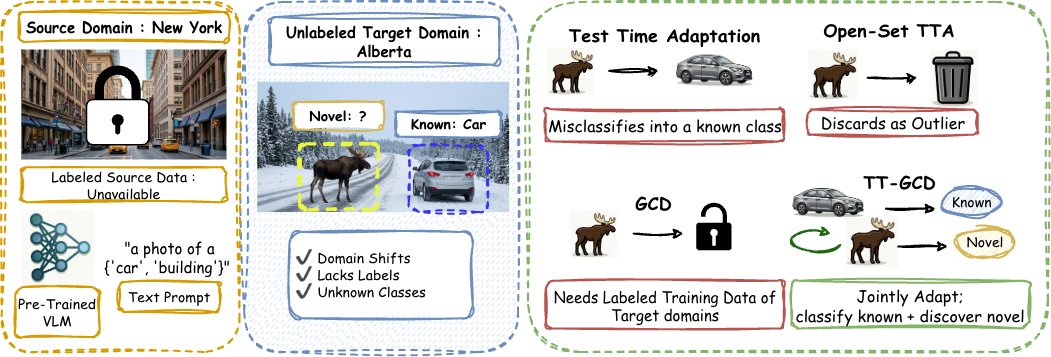}
    \caption{\textbf{The TT-GCD setting.}  A VLM trained on New York streets, with no labeled source data [\textit{Left}], is deployed on an unlabeled Alberta stream [\textit{Middle}] that combines domain shift (snow), known classes (\texttt{car}) and novel classes (\texttt{moose}). [\textit{Right}] TTA misclassifies the novel class as a known class, Open-Set TTA discards it as an outlier, GCD needs labeled target-domain data, and TT-GCD adapts to the shift, classifies the known and discovers the novel.}
    \label{fig:teaser}
    \vspace{-10pt}
\end{figure}

Vision-language models (VLMs) such as CLIP~\citep{radford2021learning} have emerged as a powerful paradigm for visual recognition, enabling generalization across a wide range of downstream vision tasks~\citep{mishra2026semanticanchortransportrobust, dong2025towards, wang2024get, caselli2026spectralgcd, schrodi2025two}. 
Yet real-world deployment poses a compound challenge as the visual distribution may shift while previously unseen categories are simultaneously encountered. Consider an autonomous driving system trained on urban streets in New York (see Fig.~\ref{fig:teaser}[\textit{Left}]) and deployed to mountain highways of Banff, Alberta (see Fig.~\ref{fig:teaser}[\textit{Middle}]).
Familiar object categories such as sedans, pedestrians, and traffic signs now appear under unfamiliar conditions of snow glare, fog, and unpaved road surfaces. Simultaneously, the system encounters entirely new object categories it was never trained on, including wildlife such as moose, deer, and bears on the roadway and recreational vehicles absent from its urban training set.
% Known categories such as vehicles and traffic signs may appear under unfamiliar snow, fog and road conditions, while at the same time, entirely novel categories such as wildlife like Moose may enter the scene.
% Yet despite their success, they still face important challenges when applied in real‑world conditions. Consider for example an autonomous driving system trained on urban streets in New York (refer Fig.~\ref{fig:teaser}[\textit{Left}]) and deployed to the mountain highways of Banff, Alberta (refer Fig.~\ref{fig:teaser}[\textit{Middle}]). The perception module faces a compound challenge. Familiar object categories such as sedans, pedestrians, and traffic signs now appear under unfamiliar conditions of snow glare, fog, and unpaved road surfaces. Simultaneously, the system encounters entirely new object categories it was never trained on, including wildlife such as moose, deer, and bears on the roadway and recreational vehicles absent from its urban training set. 
The deployed model must adapt to these domain shifts, maintain reliable recognition of known categories and also group unknown objects into coherent categories, all at inference time, without labeled data or access to the original training set.

\begin{wrapfigure}{R}{0.42\textwidth}
\centering
\includegraphics[width=0.4\textwidth]{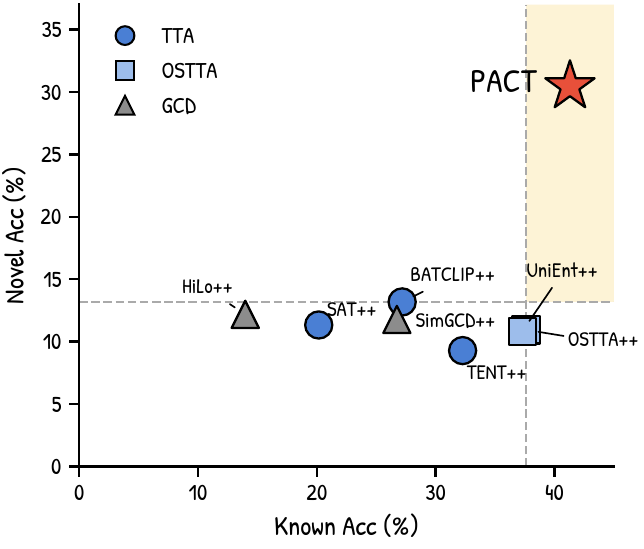}
%\caption{\textbf{Baselines sit on one axis or the other.} \colKnown{} against \colNovel{} accuracy, mean over CIFAR-10-C, CIFAR-100-C and ImageNet-C, three seeds.}
% \caption{\textbf{\texttt{Known}-class recognition and novel class discovery must be solved jointly.} 
\caption{Mean performance (\textbf{Known vs. novel}) over CIFAR-10-C, CIFAR-100-C and ImageNet-C. Dashed lines mark the best baseline on each axis.}
%Baselines built for one task stay low on \texttt{\colNovel{}} accuracy, and several also degrade \texttt{\colKnown{}} accuracy. 
% Mean over CIFAR-10-C, CIFAR-100-C and ImageNet-C.}
\label{fig:tradeoff}
\vspace{-1.4em}
\end{wrapfigure}
 
Existing paradigms (Fig.~\ref{fig:teaser}[\textit{Right}]) address only parts of this problem.
%, each under assumptions that break in the scenario above.
Test-Time Adaptation (TTA) methods~\citep{wangtent, mishra2026semanticanchortransportrobust}
handle distribution shifts by updating model parameters on unlabeled test data,  but they operate under a \emph{closed label set} assumption, leaving novel-class samples to be silently misclassified into known categories.
Generalized Category Discovery (GCD) approaches~\citep{vaze2022generalized,wen2023simgcd,chiaroni2023parametric,wang2024get} discover novel categories alongside known ones, but require \emph{labeled training data} for the known classes and assume \emph{no distribution shift} between training and inference.
HiLo~\citep{wang2025hilo} extends GCD to handle domain shifts, yet it still requires \emph{training with cross-domain labeled data}, a resource unavailable in deployment.
%% CUT: "Last,"
Additionally, Open-set TTA~\citep{gao2024unient,lee2023ostta} detects out-of-distribution samples during adaptation but \emph{rejects} them rather than organizing them into semantically coherent clusters.
% Add TTD in related work, not here since we don't baseline it
% Thus, as summarized in
% %% CUT: "table~\ref{table:settings}"
% \rev{Table}~\ref{table:settings}, no existing approach addresses all three challenges jointly under fully unsupervised test-time conditions.
Thus, these settings fail to jointly address and perform well in distribution shift, known-class recognition and novel-category discovery simultaneously under fully unsupervised test-time conditions (refer Fig.~\ref{fig:tradeoff}).
% Thus, as summarized in Table~\ref{table:settings}, no existing approach addresses all three challenges jointly under fully unsupervised test-time conditions, \textcolor{blue}{whose performance consequences on our proposed scenario are visible in Fig.~\ref{fig:tradeoff}}. %and the consequence is visible in Fig.~\ref{fig:tradeoff}. 

VLMs provide a natural foundation for bridging this gap. Their vision-language alignment enables known categories to be recognized directly from textual descriptions alone, eliminating the labeled source data required by conventional GCD.
However, an off-the-shelf VLM remains vulnerable to distribution shift and provides no direct mechanism for discovering categories absent from its text prompts. 
% VLMs offer a natural foundation for unifying these capabilities: their aligned embedding spaces allow zero-shot classification of known categories from textual descriptions alone, removing the dependency on labeled source data that limits GCD methods. Yet a VLM deployed out of the box still suffers from domain shift and has no mechanism to discover categories absent from the provided text prompts.
We therefore formulate \textbf{Test-Time Generalized Category Discovery (TT-GCD)}, where a pretrained VLM is given only the text descriptions of known categories and an unlabeled test stream containing both known and novel categories under distribution shift.
% We therefore formalize Test-Time Generalized Category Discovery (TT-GCD), as a  unified setting in which a pre-trained vision-language model with text descriptions of known categories must 
The goal is to simultaneously (1) adapt to distribution shifts, (2) classify known categories using textual supervision, and (3) discover and cluster novel categories, all at test time without any labeled data, source dataset access, or prior knowledge of how many novel categories exist.
 
%% CUT (entire HOTT method paragraph):
%% "To address TT-GCD, we propose HOTT (Hierarchical Optimal Transport for
%%  TT-GCD), a fully unsupervised framework that leverages CLIP's aligned
%%  vision-language representations. HOTT proceeds in three stages. First, an
%%  optimal transport formulation with uniform marginals computes soft
%%  assignments between test samples and text-derived known-class prototypes,
%%  and a Gaussian Mixture Model on the transport mass separates known from
%%  novel candidates. Second, known prototypes from text embeddings and novel
%%  prototypes initialized via KMeans++ are jointly refined through
%%  Sinkhorn-based Expectation-Maximization, producing a unified transport
%%  plan over all classes. Third, a lightweight adaptation stage updates only
%%  the model's LayerNorm parameters by leveraging pseudo-labels derived from
%%  the transport plan in a cross-entropy objective, ensuring that the
%%  adaptation signal is informed by both the known-class textual anchors and
%%  the discovered novel structure."
To address TT-GCD, we propose \textbf{\ours} (\textbf{P}rototype \textbf{A}ssignment for \textbf{C}ategory discovery at
\textbf{T}est time), a fully unsupervised framework with two complementary stages. First, \ours\ re-aligns shifted visual representations with the fixed text space by adapting only the LayerNorm affine parameters of the image
encoder, guided by confidence-weighted zero-shot predictions. Second, to avoid the modality mismatch between text representations for known classes and visual representations for novel ones~\citep{liang2022mind}, \ours\ represents
both as prototypes in the adapted visual space. Known and novel categories are then handled through a unified nearest-prototype assignment, with prototypes updated online as the test stream arrives. This yields a simple streaming framework in which encoder re-alignment is performed once, while subsequent adaptation operates only on category prototypes.
% To address TT-GCD, we propose \rev{\textbf{\ours}
% (\textbf{P}rototype \textbf{A}ssignment for \textbf{C}ategory discovery at
% \textbf{T}est time)}, a fully unsupervised framework that leverages CLIP's
% aligned vision-language representations. \rev{\ours\ proceeds in two stages.
% First, a lightweight adaptation stage re-aligns the shifted visual embeddings
% with the text-derived class representations, supervised by the model's own
% confident zero-shot predictions. These predictions are computed once and serve as fixed anchors during adaptation.
% %held fixed, so that adaptation does not chase its own moving outputs. 
% Second, since
% CLIP's image and text embeddings occupy distinct regions of the representation
% space~\citep{liang2022mind}, \ours\ represents known and novel categories
% homogeneously, as prototypes estimated in the visual embedding space and
% classifies every test sample through a single nearest-prototype assignment updated
% online as the stream arrives. The known/novel decision itself requires no
% separate mechanism and a sample is novel exactly when a novel prototype wins the
% assignment, and the decision remains revisable as the stream continues.}

\begin{itemize}
    \item We introduce \textbf{Test-Time Generalized Category Discovery (TT-GCD)}, a setting in which a pretrained VLM must perform domain adaptation, known-class recognition from text, and novel-category discovery
    entirely at inference time without labeled data.

    \item We propose \textbf{\ours}, a fully unsupervised framework that couples lightweight feature re-alignment with online prototype estimation, enabling a unified nearest-prototype rule to jointly recognize known
    categories and discover novel ones.

    \item We establish a unified TT-GCD evaluation protocol and adapt representative TTA, open-set TTA, and GCD methods as competitive baselines from relevant literature. Extensive experiments on five benchmarks spanning synthetic corruptions and natural domain shifts demonstrate consistent improvements over these baselines.
\end{itemize}

\section{Related Works}
\label{sec:related}

\mypar{Test-Time Adaptation (TTA).} Early methods recalibrate normalization statistics~\citep{ptbn, schneider2020} or minimize prediction entropy through normalization-layer updates~\citep{wangtent}, with later work adding sample filtering~\citep{eata} and sharpness-aware optimization~\citep{sar}. With the advent of VLMs, prompt-based~\citep{shu2022test, feng2023diversedataaugmentationdiffusions} and cache-based~\citep{Karmanov_2024_CVPR, zhang2024boostadapter} approaches emerged, alongside transport and alignment-based methods such as 
SAT~\citep{mishra2026semanticanchortransportrobust} and BATCLIP~\citep{maharana2025batclip}. All share one assumption, that the test stream contains only known categories. Novel samples are silently absorbed into known classes, a failure that closed-set evaluation cannot see.

% \begin{table}[h!]
% \caption{Comparing scenarios under different assumptions: Test-Time Adaptation (TTA), Open-Set TTA, Generalized Category Discovery (GCD), and the proposed Test-Time Generalized Category Discovery (TT-GCD).}
% \label{table:settings}
% \scriptsize
% \centering
% \begin{tabular}{lccccc}
% \toprule
% & & \textbf{Domain} & \textbf{Novel} & \textbf{Discover Novel} & \\
% \textbf{Setting} & & \textbf{Shift} & \textbf{Classes} & \textbf{Categories} & \textbf{Unsupervised} \\
% \midrule
% TTA & ~\citep{wangtent,shu2022test,maharana2025batclip,mishra2026semanticanchortransportrobust} & \cmark & \xmark & \xmark & \cmark \\
% Open-Set TTA & ~\citep{lee2023ostta,gao2024unient} & \cmark & \cmark & \xmark & \cmark \\
% GCD & ~\citep{vaze2022generalized, chiaroni2023parametric, wang2024get} & \xmark & \cmark & \cmark & \xmark \\
% GCD (\textit{w/Domain Shift}) & ~\citep{wang2025hilo} & \cmark & \cmark & \cmark & \xmark \\
% \rowcolor{gray!20} TT-GCD (\textit{Ours}) & & \cmark & \cmark & \cmark & \cmark \\
% \bottomrule
% \end{tabular}
% \end{table}

\mypar{Open-Set TTA.}
A second line of work detects out-of-distribution samples during adaptation, by entropy separation~\citep{gao2024unient}, confidence filtering~\citep{lee2023ostta} or, for VLMs, a threshold on image-text similarity~\citep{sreenivas2025rosita}, and rejects them at inference. AEO~\citep{dong2025towards} extends the entropy route to multimodal inputs. In every case the unknown is treated as noise. OWTTT~\citep{li2023owttt} is closest to our mechanism as it grows a prototype pool for out-of-distribution samples, but reports them as a single rejection class. It needs a source-trained model with source feature statistics. TT-GCD instead requires that novel samples be organized into categories, with no textual supervision.

\mypar{Generalized Category Discovery} (GCD)~\citep{vaze2022generalized} jointly classifies known and discovers novel categories from unlabeled data. Methods such as SimGCD~\citep{wen2023simgcd}, UNO~\citep{fini2021unified}, Meta-GCD~\citep{wu2023metagcd}, PIM \citep{chiaroni2023parametric} and PromptCAL~\citep{zhang2023promptcal} combine supervised learning on labeled known-class data with self-labeling, self-distillation or mutual-information objectives. VLM-based extensions ~\citep{clip-gcd, wang2024get, caselli2026spectralgcd} use CLIP's embedding space for known-novel interaction. All assume that labeled and unlabeled data share a domain, except HiLo~\citep{wang2025hilo}, which handles domain shift but needs to fine-tune the backbone on labeled source data. Every GCD method therefore depends on labeled known-class data, without which known-class boundaries degrade and clustering fails under shift.
%while HiLo~\citep{wang2025hilo} explicitly handles domain shifts through semantic-domain disentanglement. 

\mypar{The modality gap in VLMs.} Despite contrastive alignment, CLIP's image and text embeddings occupy two separate, narrow regions of the shared space, a phenomenon known as the modality gap~\citep{liang2022mind}. It arises from the cone effect at initialization, survives contrastive training~\citep{liang2022mind} and concentrates in a few embedding dimensions~\citep{schrodi2025two}. Closing it post hoc can degrade downstream performance~\citep{liang2022mind, schrodi2025two}, and closing it by retraining~\citep{eslami2025mitigate} is not available at test time, which is why \ours\ assigns among image-space prototypes (Sec.~\ref{sec:discover}).
%Despite contrastive alignment, CLIP's image and text embeddings occupy two separate, narrow regions of the shared space, the modality gap~\citep{liang2022mind}, attributed to the cone effect at initialization and preserved by contrastive optimization. Subsequent work traces the gap to a few embedding dimensions and to information imbalance between modalities~\citep{schrodi2025two}, and closes it by retraining with alignment objectives~\citep{eslami2025mitigate}. Notably, closing the gap post hoc tends to degrade downstream performance~\citep{liang2022mind, schrodi2025two}. Rather than removing the gap, \ours\ works with it: cross-modal similarities supervise adaptation, while assignment is carried out entirely among image-space prototypes.

\mypar{Test-Time Category Discovery.} On-the-fly Category Discovery~\citep{du2023ocd} and PHE~\citep{zheng2024phe} label streaming samples of known and unknown classes with hash-based descriptors, but both train on labeled known-class data and assume no distribution shift. TTD~\citep{lyu2025testtimediscoveryhashingmemory} discovers classes without training, hashing the features of a DINO backbone~\citep{dino} fine-tuned on labeled known classes, but it targets class shift alone and its purely visual hashing has no counterpart for the text anchors a VLM provides.

% \mypar{Test-Time Category Discovery.} \rev{A recent line of work discovers categories at inference time. On-the-fly Category Discovery~\citep{du2023ocd} labels streaming samples of known and unknown classes via hash-based descriptors, and PHE~\citep{zheng2024phe} improves fine-grained discovery with prototypical hash centers; both, however, learn from a labeled support set of known classes and assume no distribution shift.} TTD~\citep{lyu2025testtimediscoveryhashingmemory} performs training-free clustering via locality-sensitive hashing over visual features. However, TTD relies on a supervised source-trained backbone~\citep{dino} and labeled known-class prototypes, with hash boundaries calibrated on source domain statistics. Under domain shift, these boundaries drift and performance degrades; a failure mode TTD neither addresses nor evaluates. Furthermore, its LSH operates in a purely visual feature space, making extension to CLIP-style cross-modal embeddings non-trivial and forgoing the text-image alignment that VLMs provide for zero-shot grounding of known classes.

% \mypar{Summary:} As shown in Table~\ref{table:settings}, TTA methods handle domain drifts but not novelty, open-set TTA detects novel samples but discards them, while GCD discovers novel classes but requires labeled training data. No existing paradigm addresses all three simultaneously under fully unsupervised test-time conditions. TT-GCD is thus the first setting to introduce this deployment-realistic scenario, where \ours is designed as a single framework to address these challenges.

\section{Test-Time Generalized Category Discovery (TT-GCD)}
% \subsection{Problem Definition}
In this work, we introduce the task of \textbf{Test-Time Generalized Category Discovery (TT-GCD)}, which aligns with the practical demands of real-world deployment. In this setting, we use only a pretrained VLM and an unlabeled test set, to assign semantic labels to \emph{all} test samples covering both known and novel categories while simultaneously adapting to distribution shift. 
%using only a pretrained vision-language model and an unlabeled test set.

Formally, we assume access to a pretrained VLM %vision-language model 
(e.g., CLIP~\citep{radford2021learning}) consisting of an image encoder $f_I: \mathcal{X} \rightarrow \mathbb{R}^d$ and a text encoder $f_T: \mathcal{T} \rightarrow \mathbb{R}^d$. At inference, the model receives an unlabeled test set $\mathcal{D}_{\text{test}} = \{\vx_n\}_{n=1}^{N}$, where each sample is observed once. The test stream contains samples from known categories $\Cknown$ and novel categories $\Cnovel$, with $\Cknown \cap \Cnovel = \emptyset$ and is also subjected to distribution shift.

We denote the $\ell_2$-normalized visual embedding of test image $\vx_n$ by $\vv_n=f_I(\vx_n)\in\mathbb{R}^d$, and the $\ell_2$-normalized text embedding of known class $k$ by $\vz_k=f_T(t_k)\in\mathbb{R}^d$, where $t_k$ is a text prompt for class $k$, e.g., $\textsc{\textcolor{gray}{``A photo of a [class$_k$]''}}$. Following standard CLIP-based TTA settings~\citep{shu2022test,maharana2025batclip,mishra2026semanticanchortransportrobust}, the names of the $K=|\Cknown|$ known categories are available at inference, providing semantic grounding through their text embeddings. In contrast, novel categories have no textual descriptions and must be discovered directly from the unlabeled test stream. The objective of TT-GCD is to assign every test sample either to one of the $K$ known categories or to one of the $L=|\Cnovel|$ novel categories, while adapting to the distribution shift. Also, to better match real-world deployment, $L$ is unknown at test time and must be inferred from the stream. For fair comparison with baseline methods, the main protocol supplies it to every method for comparability, and Sec. ~\ref{lest} reports results when it is estimated from the stream.
TT-GCD therefore requires two capabilities: (1) \textit{known-category classification}, using the available class descriptions, and (2) \textit{novel-category discovery}, by grouping novel samples into semantically coherent categories. Both must be performed under distribution shift, using only the pretrained VLM and the unlabeled test stream. As summarized in Table~\ref{table:settings}, TT-GCD unifies domain adaptation,
known-category recognition and novel-category discovery within a fully unsupervised test-time setting.

\begin{table*}[!h]
\centering
\caption{Comparison of TT-GCD with related test-time and category-discovery
settings. TT-GCD uniquely combines distribution shift, novel categories,
novel-category discovery, and fully unlabeled test-time adaptation.}
\label{table:settings}
\scriptsize
\setlength{\tabcolsep}{3.5pt}
\renewcommand{\arraystretch}{1.15}
\begin{tabular}{lcccc}
\toprule
\textbf{Setting} &
\shortstack{\textbf{Domain}\\\textbf{Shift}} &
\shortstack{\textbf{Novel}\\\textbf{Classes}} &
\shortstack{\textbf{Novel}\\\textbf{Discovery}} &
\shortstack{\textbf{No Labeled}\\\textbf{Train Data}} \\
\midrule
TTA~\citep{wangtent,shu2022test,maharana2025batclip,
mishra2026semanticanchortransportrobust}
    & \cmark & \xmark & \xmark & \cmark \\

Open-Set TTA~\citep{lee2023ostta,gao2024unient}
    & \cmark & \cmark & \xmark & \cmark \\

GCD~\citep{vaze2022generalized,chiaroni2023parametric,wang2024get}
    & \xmark & \cmark & \cmark & \xmark \\

GCD w/ Shift~\citep{wang2025hilo}
    & \cmark & \cmark & \cmark & \xmark \\

\rowcolor{gray!20}
\textbf{TT-GCD (Ours)}
    & \cmark & \cmark & \cmark & \cmark \\
\bottomrule
\end{tabular}
\end{table*}

\section{Prototype Assignment for Category Discovery at Test Time (\ours)}
\label{sec:method}
\ours\ decouples TT-GCD into two problems, each solved by minimizing an explicit objective. \textbf{(1) Feature Re-Alignment.} At test time, distribution shift primarily affects the visual input, while the text encoder continues to receive the same known class-name prompts. As a result, the shifted visual embeddings may become misaligned with their corresponding text embeddings. 
We first re-align the visual representations with the fixed text features by adapting only the LayerNorm affine parameters of the image encoder, providing a lightweight mechanism for test-time adaptation~\citep{schneider2020,wangtent, maharana2025batclip} (Sec.~\ref{sec:align}).
\textbf{(2) Unified Category Discovery.} In the re-aligned space, known-class recognition and novel-category discovery are formulated jointly through prototype assignment on the unit hypersphere, with both known and novel categories represented by prototypes in the visual embedding space (Sec.~\ref{sec:discover}).

% the distribution shift enters through the images: the text encoder receives the same class-name prompts as at pretraining, while $f_I$ receives input images from the shifted distribution. The visual embeddings $\vv_n$ are therefore displaced relative to the text embeddings $\vz_k$. We first re-align the two by adapting the LayerNorm affine parameters of both encoders, where domain-specific variation concentrates~\citep{schneider2020, wangtent, maharana2025batclip}(Sec.~\ref{sec:align}). 
% \textbf{(2) Unified Category Discovery.} Next, in the
% corrected space, classification and discovery become a single clustering problem on the unit sphere, with every category, known or novel, represented by a prototype in the visual embedding space (Sec.~\ref{sec:discover}).

\subsection{Feature Re-Alignment}
\label{sec:align}
 
\paragraph{Initial buffer.}
Let $\theta_I$ denote the LayerNorm affine parameters of the image encoder with pretrained values $\theta^0_I$. We write the corresponding visual embedding as $\vv_n^{\theta_I}$, while the text embeddings $\{\vz_k\}_{k=1}^{K}$ remain fixed. The zero-shot probability over the $K$ known classes is

\begin{equation}
  p_n^{\theta_I}(k)
  =
  \frac{
  \exp\!\left(\vv_n^{\theta_I\top}\vz_k/\tau\right)
  }{
  \sum_{j=1}^{K}
  \exp\!\left(\vv_n^{\theta_I\top}\vz_j/\tau\right)
  },
  \label{eq:zsp}
\end{equation}
where $\tau$ is the temperature.
Before adaptation, we collect the first $W$ test samples in an initial buffer $\mathcal{B}$. For each $\vx_n\in\mathcal{B}$, the pretrained image encoder $\theta^0_I$ together with the fixed text embeddings provides a pseudo-label $\hat{y}_n$ and confidence weight $\omega_n$:
\begin{equation}
  \hat{y}_n
  =
  \arg\max_{k} p_n^{\theta^0_I}(k),
  \qquad
  \omega_n
  =
  p_n^{(1)}-p_n^{(2)}
  \in[0,1],
  \label{eq:weight}
\end{equation}
where $p_n^{(1)}$ and $p_n^{(2)}$ denote the largest and second-largest zero-shot probabilities under $\theta^0_I$, respectively. The resulting $\hat{y}_n$ and $\omega_n$ are computed once and remain fixed throughout
re-alignment.

% Before adaptation, we store the first $W$ samples of the test stream in a buffer $\mathcal{B}$ and for each $\vx_n\in\mathcal{B}$, we compute the zero-shot prediction, $\hat{y}_n$ and a confidence weight, $\omega_n$, from the initial model,
% \begin{equation}
%   \hat{y}_n
%   \;=\;
%   \arg\max_{k}\; p_n^{\theta_0}(k),
%   \qquad
%   \omega_n
%   \;=\;
%   p_n^{(1)} - p_n^{(2)} \;\in\; [0,1],
%   \label{eq:weight}
% \end{equation}
% where $\theta^0$ denote the initial, pre-trained values of the LayerNorm affine parameters, i.e., $\theta^0=(\theta_I^0,\theta_T^0)$, and $p_n^{(1)}$ and $p_n^{(2)}$ are the highest and second-highest values of $p_n^{\theta_0}$. Both $p_n^{(1)}$ and $p_n^{(2)}$ are computed once, at $\theta_0$, and remain fixed during adaptation.

\paragraph{Re-alignment objective.}

We re-align the shifted visual representations with the fixed text space by optimizing only the LayerNorm affine parameters $\theta_I$ of the image encoder. Specifically,
\begin{equation}
  \theta_I^{\star}
  =
  \arg\min_{\theta_I}
  -\frac{1}{\sum_{n\in\mathcal{B}}\omega_n}
  \sum_{n\in\mathcal{B}}
  \omega_n
  \log p_n^{\theta_I}(\hat{y}_n).
  \label{eq:loss_theta}
\end{equation}

Weighting samples by $\omega_n$ emphasizes predictions that are more reliable under the initial model. Expanding the cross-entropy term gives,
\begin{equation}
  -\log p_n^{\theta_I}(\hat{y}_n)
  =
  \underbrace{
  -\frac{1}{\tau}
  \vv_n^{\theta_I\top}
  \vz_{\hat{y}_n}
  }_{\text{alignment}}
  +
  \underbrace{
  \log\!\sum_{j=1}^{K}
  \exp\!\left(
  \vv_n^{\theta_I\top}\vz_j/\tau
  \right)
  }_{\text{normalization}}.
  \label{eq:ce-decomp}
\end{equation}

The first term pulls each visual embedding toward the text embedding of its assigned class, while the second accounts for the competing known classes. Together, they increase the relative alignment with the assigned class, re-aligning the shifted visual representations with the initial text space. Importantly, while $\theta_I$ is optimized, the text embeddings $\{\vz_k\}$, pseudo-labels $\hat{y}_n$, and confidence weights $\omega_n$ remain fixed. The fixed text embeddings provide semantic anchors, while confidence-weighted pseudo-labels reduce the influence of unreliable predictions from the evolving visual encoder mitigating confirmation bias~\citep{arazo2020pseudo}.
% It can be seen as a confidence‑weighted self‑alignment step, using the initial model as a fixed anchor.

\subsection{Unified Category Discovery}
\label{sec:discover}

After re-alignment, we freeze $\theta_I^{\star}$ and denote the resulting adapted visual embeddings by $\vv_n^\star$. 
%The text embeddings $\{\vz_k\}_{k=1}^{K}$ remain fixed throughout. 
All subsequent category assignments are performed in this adapted space.

% After re-alignment, the adapted parameters $\theta^{\star}$ are frozen. Each image is subsequently encoded as $\vv_n^{\star}=f_I^{\theta^{\star}}(\vx_n)$, and the text embeddings $\{\vz_k^{\star}\}$ are recomputed once with the adapted text encoder. All subsequent predictions use these adapted embeddings.

\paragraph{Unified feature space.} Although CLIP aligns image and text representations, a modality gap remains between the two embedding spaces, leading to systematically different image--text and image--image similarities~\citep{liang2022mind}. Directly comparing text embeddings for known classes against visual prototypes for novel classes can therefore bias category assignment toward one modality. We avoid this by representing both known and novel categories as visual prototypes in the unified space.

Since the adapted visual features are $\ell_2$-normalized, they lie on the unit hypersphere. We model them as a mixture of von Mises--Fisher (vMF) components with shared concentration, one per category~\citep{banerjee2005vmf,mardia2000directional}. Under hard assignments and uniform priors, maximum-likelihood estimation of this model is equivalent to spherical $k$-means~\citep{banerjee2005vmf}. This motivates our prototype
construction: known prototypes are estimated from confidence-weighted zero-shot assignments, while novel prototypes are initialized and subsequently refined from their assigned visual samples.

\paragraph{Known prototypes.}
For each known class $k$, we aggregate the adapted embeddings in the initial buffer of those samples that receive predicted class $\hat{y}_n = k$, weighted by their zero-shot confidence $\omega_n$, and project the resulting vector onto the unit sphere:
\begin{equation}
  \bar{\boldsymbol{\mu}}_k
  =
  \sum_{\substack{n\in\mathcal{B}\\ \hat{y}_n=k}}
  \omega_n\,\vv_n^{\star},
  \qquad
  \boldsymbol{\mu}_k
  =
  \frac{\bar{\boldsymbol{\mu}}_k}
       {\big\lVert\bar{\boldsymbol{\mu}}_k\big\rVert_2},
  \qquad
  e_k
  =
  \sum_{\substack{n\in\mathcal{B}\\ \hat{y}_n=k}}
  \omega_n,
  \label{eq:known-proto}
\end{equation}
% so that $\boldsymbol{\mu}_k$ is the confidence-weighted mean direction of class $k$ and $e_k$ the total confidence the class received. Prototypes estimated from a few low-confidence pseudo-labels are unreliable, so we retain only the classes with $e_k \ge e_{\min}$, forming the active known set $\mathcal{A}$.
Thus, $\boldsymbol{\mu}_k$ forms the confidence-weighted estimated prototypes of class $k$ 
%is the confidence-weighted mean direction of class $k$ 
and $e_k$ the total confidence of that predicted class. %k$. 
To avoid unreliable estimates supported by few or low-confidence samples, we retain only classes with total confidence higher than a given threshold $e_{\min}$. These retained known prototypes form the active known set $\mathcal{A}
=
\left\{\boldsymbol{\mu}_k:e_k\ge e_{\min}
\right\}$.

\paragraph{Novel prototypes.}
The initial buffer may also contain samples from novel categories, for which neither class names nor text embeddings are available. Unlike known classes, their prototypes therefore cannot be estimated from zero-shot assignments and must instead be discovered from the geometry of the adapted visual features. We treat the active known prototypes $\mathcal{A}$ as pre-existing centers and sequentially initialize $L$ additional prototypes $\{\boldsymbol{\nu}_l\}_{l=1}^{L}$ using spherical $k$-means++~\citep{arthur2007}. Before selecting the $l$-th novel prototype, let
$  \mathcal{C}_l
  =
  \mathcal{A}
  \cup
  \{\boldsymbol{\nu}_r\}_{r<l}
  \label{eq:seedset}$
denote the set of prototypes available so far. For each buffered feature $\vv_n^\star$, we compute its squared distance to the closest existing prototype as
\begin{equation}
  D_l^2(n)
  =
  \min_{c\in\mathcal{C}_l}
  \left\lVert\vv_n^{\star}-c\right\rVert_2^2
  =
  2\Big(1-\max_{c\in\mathcal{C}_l}\vv_n^{\star\top}c\Big),
  \label{eq:d2}
\end{equation}
Thus, features already well represented by an active known prototype or a previously selected novel prototype receive low sampling probability, whereas features in uncovered regions receive higher probability. Repeating this process yields $L$ novel prototypes that complement the active known prototypes and provide initial coverage of the buffered visual feature space.

\paragraph{Class assignment.}
After initialization, the unified prototype set is
\begin{equation}
  \mathcal{P}
  =
  \mathcal{A}
  \cup
  \{\boldsymbol{\nu}_l\}_{l=1}^{L},
  \label{eq:protoset}
\end{equation}
where known and novel categories are represented in the same unified visual embedding space. Under the vMF model with shared concentration and uniform priors, MAP based inference reduces to nearest-prototype assignment by
cosine similarity~\citep{banerjee2005vmf}:
\begin{equation}
  \ell(\vv^\star)
  =
  \arg\max_{c\in\mathcal{P}}
  \vv^{\star\top}c .
  \label{eq:assign}
\end{equation}
An assignment to $\boldsymbol{\mu}_k$ predicts known class $k$, whereas an assignment to $\boldsymbol{\nu}_l$ identifies discovered novel category $l$.

\paragraph{Streaming adaptation.}
Re-alignment is performed only once on the initial buffer $\mathcal{B}$. Thereafter, $\theta_I^{\star}$ remains frozen and adaptation proceeds by updating only the category prototypes as new test batches arrive. For each incoming batch $\mathcal{B}_t$, all assignments are computed before updating the prototypes. Known categories remain semantically anchored by the adapted zero-shot classifier: let $\hat{y}_n^\star$ and $\omega_n^\star$ denote the
prediction and confidence obtained from Eq.~\ref{eq:zsp} using $\theta_I^{\star}$. Novel categories, for which no textual supervision is available, instead follow the unified prototype assignment in Eq.~\ref{eq:assign}. We therefore
define
$  \mathcal{K}_k^t
  =
  \{n\in\mathcal{B}_t:\hat{y}_n^\star=k\}
$ and $
  \mathcal{I}_l^t
  =
  \{n\in\mathcal{B}_t:
  \ell(\vv_n^\star)=\boldsymbol{\nu}_l\}$
as the samples supporting known class $k$ and novel category $l$,
respectively.
For each known class, we compute the confidence-weighted mean direction and update its prototype
\begin{equation}
  \boldsymbol{\mu}_k
  \leftarrow
  \nrm{(1-\eta)\,\boldsymbol{\mu}_k+\eta\,\tilde{\boldsymbol{\mu}}_k^{\,t}},
  \qquad
  \tilde{\boldsymbol{\mu}}_k^{\,t}
  =
  \nrm{\sum_{n\in\mathcal{K}_k^t}\omega_n^{\star}\,\vv_n^{\star}},
  \label{eq:known-update}
  \end{equation}
  Similarly, novel prototypes are updated as
  \begin{equation}
  \boldsymbol{\nu}_l
  \leftarrow
  \nrm{(1-\lambda)\,\boldsymbol{\nu}_l+\lambda\,\tilde{\boldsymbol{\nu}}_l^{\,t}},
  \qquad
  \tilde{\boldsymbol{\nu}}_l^{\,t}
  =
  \nrm{\sum_{n\in\mathcal{I}_l^t}\vv_n^{\star}},
  \label{eq:novel-update}
  \end{equation}
  where $\eta,\lambda\in(0,1]$ control the update rates for known and novel prototypes, respectively. If no samples support a prototype in $\mathcal{B}_t$, it remains unchanged.
  Finally, confidence for each known class accumulates over the stream:
\begin{equation}
  e_k
  \leftarrow
  e_k
  +
  \sum_{n\in\mathcal{K}_k^t}
  \omega_n^\star,
  \label{eq:evidence}
\end{equation}
and its prototype enters the active set $\mathcal{A}$ once $e_k\ge e_{\min}$.

\section{Experiments}
\subsection{Setup}
\label{sec:setup}
\mypar{Datasets.}
We evaluate \ours on five benchmarks spanning synthetic corruption and natural domain shift. We use the standard TTA benchmarks CIFAR-10-C, CIFAR-100-C and ImageNet-C~\citep{hendrycks2021natural}. We also evaluate on CUB-C~\citep{wang2025hilo}, which applies seven corruptions to the fine-grained CUB dataset~\citep{WahCUB_200_2011} and DomainNet~\citep{peng2019momentmatchingmultisourcedomain}, comprising 345 classes across six domains. For each dataset, a proportion $\rho$ of classes is designated as known and the remaining $(1-\rho)$ as novel, we use $\rho=0.5$, i.e., 50\% known and 50\% novel classes. Class splits and seeds are detailed in App.~\ref{app:data}. 

\mypar{Baselines.}
As no prior method directly addresses TT-GCD, we adapt seven representative baselines to this setting. For TTA methods, TENT++~\citep{wangtent}, BATCLIP++~\citep{maharana2025batclip}, SAT++~\citep{mishra2026semanticanchortransportrobust}, OSTTA++~\citep{lee2023ostta}, and UniEnt++~\citep{gao2024unient}, we first separate known and novel samples using a two-component Gaussian mixture over their maximum similarity to the known text embeddings. Adaptation follows each original method on the predicted known subset, updating only LayerNorm parameters. Known samples use the adapted model predictions, while novel samples are grouped into $L$ clusters using $k$-means. For GCD methods, SimGCD++~\citep{wen2023simgcd} and HiLo++~\citep{wang2025hilo}, we replace their required labeled samples with zero-shot pseudo-labeled known samples and keep the backbone fixed. Their learned prototypes are used for both known and novel categories. Full adaptation details are in App.~\ref{app:adapt}.

\mypar{Evaluation protocol.} We report clustering accuracy (\%) following the standard practice of prior GCD literature ~\citep{wen2023simgcd} on known classes (\colKnown), novel classes (\colNovel) and all classes (\texttt{All}), read from a single Hungarian assignment over the full stream. Results are averaged over all corruption types or domains of each benchmark. Tables~\ref{tab:main} and~\ref{tab:domainnet} report the mean and standard deviation over three seeds. More details in App.~\ref{app:data}. 

\mypar{Implementation details.}
We use CLIP ViT-B/16 as the backbone unless stated otherwise, with the prompt \texttt{a photo of a \{\}.} and batch size 128. Re-alignment is performed once per stream with a learning rate of $10^{-2}$, while known and novel prototypes are updated with rates $\eta=0.05$ and $\lambda=0.10$, respectively. All hyperparameters for \ours\ and the baselines are selected on four held-out CIFAR-100-C corruptions and kept fixed across benchmarks(App.~\ref{app:impl}).

\subsection{Main results}
%We first assess performance under synthetic corruptions, and then turn to natural shift and fine-grained categories.

\mypar{Corrupted datasets.} The results on the three corruption benchmarks are reported in Table~\ref{tab:main}. \ours\ obtains the best \texttt{All} accuracy on the three, \textbf{outperforming the second best method by 16.8} points on CIFAR-10-C, \textbf{9.3} on CIFAR-100-C and \textbf{5.3} on ImageNet-C. Furthermore, it yields the best accuracy in eight of the nine \texttt{All}, \texttt{\colKnown{}} and \texttt{\colNovel{}} columns. A closer look reveals a consistent pattern, i.e., no baseline is competitive on both halves of the label space. For example, on CIFAR-10-C the three highest \texttt{\colKnown{}} accuracies, UniEnt++ ($71.18$), OSTTA++ ($70.43$) and TENT++ ($60.06$), coincide with the three weakest \texttt{\colNovel{}} results ($7.77$, $8.06$ and $6.75$). %These baselines rely on objectives that sharpen the known-class assignment. 
HiLo++ shows the reverse, with the best baseline \texttt{\colNovel{}} on CIFAR-10-C ($16.41$) and the worst \texttt{All} on CIFAR-100-C ($5.69$), as its robustness relies on cross-domain labeled training, which TT-GCD does not provide (App.~\ref{app:adapt}). 

\begin{table*}[t]
\centering
\caption{\textbf{TT-GCD on corruption benchmarks} (severity 5, mean over 15 corruption
types). Clustering accuracy (\%), mean over 3 seeds with standard deviation. Best in \textbf{bold}, second best \underline{underlined}.}
\label{tab:main}
\fontsize{7.5}{9}\selectfont
\setlength{\tabcolsep}{2pt}
\begin{tabular}{@{}l ccc ccc ccc@{}}
\toprule
& \multicolumn{3}{c}{\textbf{CIFAR-10-C}} & \multicolumn{3}{c}{\textbf{CIFAR-100-C}}
& \multicolumn{3}{c}{\textbf{ImageNet-C}}\\
\cmidrule(lr){2-4}\cmidrule(lr){5-7}\cmidrule(lr){8-10}
Method & All & \colKnown & \colNovel & All & \colKnown & \colNovel
       & All & \colKnown & \colNovel \\
\midrule
\multicolumn{10}{@{}l}{\textbf{Test Time Adaptation}}\\
\midrule
TENT++ & \vc{33.41}{.10} & \vc{60.06}{.12} & \vc{6.75}{.25}
       & \vc{15.78}{.01} & \vc{17.16}{.02} & \vc{14.41}{.04}
       & \vc{13.14}{.07} & \vc{19.58}{.12} & \vc{6.69}{.03}\\
BATCLIP++ & \vc{29.24}{.08} & \vc{46.88}{.01} & \vc{11.59}{.16}
       & \vc{15.57}{.01} & \vc{15.31}{.07} & \vc{15.83}{.09}
       & \vc{15.74}{.06} & \vc{18.83}{.05} & \vc{12.64}{.07}\\
SAT++ & \vc{23.43}{.07} & \vc{34.48}{.31} & \vc{12.39}{.29}
       & \vc{13.25}{.07} & \vc{13.69}{.18} & \vc{12.81}{.12}
       & \vc{10.55}{.03} & \vc{12.31}{.04} & \vc{8.80}{.03}\\
\addlinespace[2pt]
\midrule
\multicolumn{10}{@{}l}{\textbf{Open-Set Test Time Adaptation}}\\
\midrule
OSTTA++ & \vc{39.24}{.11} & \vc{\underline{70.43}}{.31} & \vc{8.06}{.12}
       & \vc{19.29}{.14} & \vc{20.97}{.19} & \vc{17.61}{.08}
       & \vc{14.30}{.06} & \vc{\underline{21.44}}{.11} & \vc{7.16}{.10}\\
UniEnt++ & \vc{\underline{39.48}}{.13} & \vc{\textbf{71.18}}{.55} & \vc{7.77}{.74}
       & \vc{\underline{19.79}}{.09} & \vc{\underline{21.70}}{.17}
       & \vc{\underline{17.88}}{.04}
       & \vc{12.91}{.05} & \vc{19.00}{.16} & \vc{6.83}{.07}\\
\addlinespace[2pt]
\midrule
\multicolumn{10}{@{}l}{\textbf{Generalized Category Discovery}}\\
\midrule
SimGCD++ & \vc{28.49}{1.02} & \vc{47.50}{1.66} & \vc{9.48}{1.21}
       & \vc{15.20}{.37} & \vc{14.77}{.31} & \vc{15.64}{.54}
       & \vc{14.07}{.28} & \vc{17.93}{.39} & \vc{10.21}{.17}\\
HiLo++ & \vc{16.63}{.16} & \vc{16.85}{1.42} & \vc{\underline{16.41}}{1.67}
       & \vc{5.69}{.06} & \vc{4.89}{.15} & \vc{6.49}{.21}
       & \vc{\underline{16.93}}{.60} & \vc{20.19}{.71} & \vc{\underline{13.66}}{.13}\\
\midrule
\rowcolor{gray!12}
\ours\ (\textit{Ours}) & \vc{\textbf{56.31}}{.58} & \vc{65.39}{.60} & \vc{\textbf{47.24}}{1.09}
       & \vc{\textbf{29.11}}{.11} & \vc{\textbf{33.67}}{.18} & \vc{\textbf{24.55}}{.28}
       & \vc{\textbf{22.20}}{.12} & \vc{\textbf{24.84}}{.27} & \vc{\textbf{19.56}}{.03}\\
\bottomrule
\end{tabular}
\end{table*}

\begin{table*}[h!]
\centering
\caption{\textbf{TT-GCD under natural shift.} DomainNet, per domain, $K{=}172$. Clustering accuracy (\texttt{All}, \%), mean over 3 seeds with the standard deviation. Best in \textbf{bold}, second best \underline{underlined}. \colKnown{} and \colNovel{} per domain are reported in App.~\ref{app:dn}.}
\label{tab:domainnet}
\fontsize{9.5}{11}\selectfont
\setlength{\tabcolsep}{4pt}
\begin{tabular}{@{}l cccccc c@{}}
\toprule
Method & Clipart & Infograph & Painting & Quickdraw & Real & Sketch & \textbf{Average} \\
\midrule
TENT++    & \vc{28.33}{.19} & \vc{17.42}{.09} & \vc{26.95}{.04} & \vc{0.52}{.00}
          & \vc{\underline{31.69}}{.03} & \vc{23.58}{.14} & 21.41 \\
BATCLIP++ & \vc{\underline{29.09}}{.15} & \vc{16.64}{.14} & \vc{\underline{28.10}}{.31}
          & \vc{\underline{6.04}}{.09} & \vc{30.63}{.39} & \vc{\underline{24.85}}{.40} & \underline{22.56} \\
SAT++     & \vc{14.17}{.28} & \vc{14.28}{.41} & \vc{9.81}{.48} & \vc{2.20}{.01}
          & \vc{7.93}{.14} & \vc{10.90}{.09} & 9.88 \\
OSTTA++   & \vc{26.82}{.34} & \vc{19.19}{.18} & \vc{25.69}{.40} & \vc{5.71}{.21}
          & \vc{27.21}{.67} & \vc{23.92}{.12} & 21.42 \\
UniEnt++  & \vc{28.01}{.09} & \vc{\underline{19.70}}{.14} & \vc{26.81}{.17} & \vc{5.50}{.25}
          & \vc{30.24}{.34} & \vc{24.80}{.37} & 22.51 \\
SimGCD++  & \vc{17.10}{15.65} & \vc{10.53}{4.56} & \vc{15.35}{16.43} & \vc{5.23}{.02}
          & \vc{17.97}{14.91} & \vc{8.43}{2.19} & 12.44 \\
HiLo++    & \vc{6.13}{.38} & \vc{5.63}{.51} & \vc{6.76}{.18} & \vc{2.66}{.26}
          & \vc{5.94}{.20} & \vc{5.11}{.08} & 5.37 \\
\midrule
\rowcolor{gray!12}
\ours\ (\textit{Ours}) & \vc{\textbf{57.22}}{1.90} & \vc{\textbf{34.50}}{.75} & \vc{\textbf{54.49}}{.93}
          & \vc{\textbf{20.74}}{.45} & \vc{\textbf{65.04}}{.83} & \vc{\textbf{49.36}}{.59} & \textbf{46.89} \\
\bottomrule
\end{tabular}
\end{table*}

\mypar{Natural shift (Table~\ref{tab:domainnet}).} In this scenario, we evaluate \ours changes in visual style shifts across domains. We observe that \textbf{\ours\ still remains the best method on all six domains} in DomainNet, with a mean accuracy of $\boldsymbol{46.89}$ \textbf{\textit{versus}} $\boldsymbol{22.56}$ for the strongest baseline, and per-domain margins ranging from $\boldsymbol{14.7}$ (Quickdraw) to $\boldsymbol{33.4}$ (Real). Among the baselines, it is noteworthy to mention that TENT++ collapses to $0.52$ on Quickdraw, a domain that is particularly challenging for all baselines, where the best baseline yields 6.04 of accuracy. In contrast, \ours remains robust and achieves 20.74, representing a substantial margin over every other approach. 

\begin{wraptable}[11]{r}{0.37\textwidth}
\vspace{-2.1\intextsep}
\centering
\caption{\textbf{Fine-grained shift on CUB-C.} \texttt{All} (\%), severity 5.}
\label{tab:cub}
\fontsize{7.5}{8.6}\selectfont
\setlength{\tabcolsep}{2.5pt}
\begin{tabular}{@{}lccc@{}}
\toprule
Method & B/16 & L/14 & BioCLIP-2 \\
\midrule
TENT++    & 9.54  & 15.01 & \underline{31.97} \\
BATCLIP++ & 11.57 & 15.52 & 28.63 \\
SAT++     & 9.11  & 10.42 & 26.22 \\
OSTTA++   & 8.86  & 14.00 & 28.82 \\
UniEnt++  & 7.57  & 13.96 & 20.58 \\
SimGCD++  & \underline{13.74} & \underline{17.85} & 24.76 \\
HiLo++    & 5.83  & 7.08  & 9.57 \\
\midrule
\rowcolor{gray!12}
\ours\ (\textit{Ours}) & \textbf{18.20} & \textbf{23.59} & \textbf{45.48} \\
\bottomrule
\end{tabular}
\vspace{-1.0\intextsep}
\end{wraptable}

\mypar{Fine-grained categories.} Finally, Table~\ref{tab:cub} reports the results on CUB-C with three backbones. The accuracy of every method drops considerably with respect to the coarse-grained benchmarks. Yet, \textbf{\ours\ yields the best \texttt{All} accuracy with the three backbones} ($18.20$, $23.59$ and $45.48$). Moving from CLIP B/16 to BioCLIP-2 improves all methods, e.g., TENT++ gains $22.4$ points, which indicates that a large part of the fine-grained gap lies in the representation. However, the gain of TENT++ comes entirely from \texttt{\colKnown{}}, whereas the \texttt{\colNovel{}} accuracy achieved by \ours\ more than doubles, from $15.09$ to $34.32$ (App.~\ref{app:cub}).

\subsection{Ablation studies}
\paragraph{Contribution of each stage.} 
Table~\ref{tab:stage2x2} jointly ablates the two key components of \ours: feature re-alignment(Sec. ~\ref{sec:align}) and visual prototypes for known classes in unified category discovery(Sec.\ref{sec:discover}).
Without re-alignment, the encoder remains at $\theta_I^0$ and without visual prototypes, known classes are represented by their text embeddings. Both components contribute to substantial performance gains. With visual prototypes fixed, re-alignment improves \texttt{All} accuracy by $5.91$ points on CIFAR-100-C and $4.51$ on DomainNet. Conversely, with re-alignment enabled, replacing text embeddings with visual prototypes improves \texttt{All} by $6.85$ and $12.53$ points, respectively. Their strongest effects are complementary: visual prototypes improve \texttt{\colKnown{}} by $11.70$--$14.55$ points on CIFAR-100-C and $20.80$--$27.28$ on DomainNet, supporting the modality-gap motivation in Sec.~\ref{sec:discover}. Re-alignment consistently improves \texttt{\colNovel{}} by $5.84$--$9.34$ points despite using only known-class pseudo-labels, indicating that adapting the shared visual encoder also benefits novel categories.
\begin{table*}[h!]
% \vspace{\intextsep}
\centering
\caption{\textbf{Component ablation.} Clustering accuracy (\%)}
\label{tab:stage2x2}
\fontsize{8.5}{10}\selectfont
\setlength{\tabcolsep}{4.0pt}
\begin{tabular}{@{}cc ccc ccc@{}}
\toprule
& & \multicolumn{3}{c}{\textbf{CIFAR-100-C}} & \multicolumn{3}{c}{\textbf{DomainNet}} \\
\cmidrule(lr){3-5} \cmidrule(l){6-8}
Re-align & Known proto & All & \colKnown & \colNovel & All & \colKnown & \colNovel \\
\midrule
--     & Text  & 15.11 & 13.83 & 16.39 & 25.22 & 24.45 & 25.91 \\
--     & Image & 23.08 & 28.38 & 17.77 & 42.13 & 51.73 & 32.62 \\
\cmark & Text  & 22.14 & 22.04 & 22.23 & 34.11 & 32.90 & 35.25 \\
\rowcolor{gray!12}
\cmark & Image & \textbf{28.99} & \textbf{33.74} & \textbf{24.25} & \textbf{46.64} & \textbf{53.70} & \textbf{39.68} \\
\bottomrule
\end{tabular}
\end{table*}

\begin{figure*}[t]
\centering
\begin{minipage}[b]{0.49\linewidth}
\centering
\includegraphics[height=1.55in]{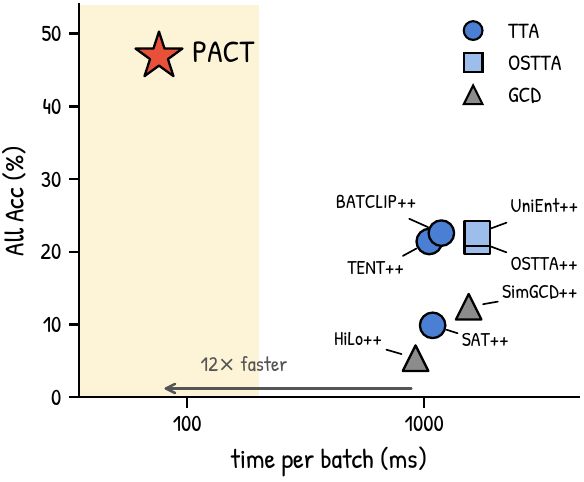}\\[-2pt]
{\small (a) Streaming efficiency}
\end{minipage}\hfill
\begin{minipage}[b]{0.49\linewidth}
\centering
\includegraphics[height=1.55in]{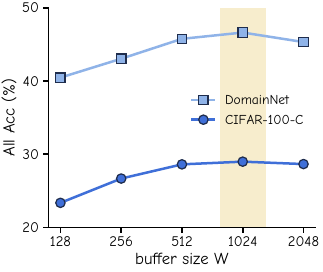}\\[-2pt]
{\small (b) Buffer size $W$}
\end{minipage}
\caption{(a) Time per batch of 128 (log scale) against mean \texttt{All} on DomainNet. (b) \texttt{All} (\%) as a function of the buffer size $W$.}
\label{fig:streaming}
% \vspace{-\intextsep}
\end{figure*}

\mypar{Streaming efficiency.}
Once re-alignment is complete, \ours\ freezes the encoder and updates only the prototypes, requiring no backward pass during streaming. As shown in Fig. ~\ref{fig:streaming}(a), \ours\ processes a batch of 128 in $76$\,ms, compared with $919$--$1675$\,ms for online adaptation baselines, yielding a $12$--$22\times$ speedup. The one-time re-alignment costs $30$\,s per stream (App.~\ref{app:cost}).

\mypar{Buffer size.} Fig. ~\ref{fig:streaming}(b) varies the buffer size $W$. \texttt{All} increases monotonically up to $W{=}1024$ on both benchmarks, from $23.36$ to $28.99$ on CIFAR-100-C and from $40.48$ to $46.64$ on DomainNet, and drops at $W{=}2048$ by $0.3$ and $1.3$. The buffer samples receive their known-class prediction from $\theta_0$ (Eq.~\ref{eq:zsp}), so a larger buffer improves the re-alignment and the prototype estimates while placing a larger fraction of the stream under the pretrained model.
% \begin{table*}[h]
% \centering
% \caption{\textbf{Buffer size $W$.} \texttt{All} (\%).}
% \label{tab:abl-W}
% \fontsize{8.5}{11}\selectfont
% \setlength{\tabcolsep}{3.5pt}
% \begin{tabular}{lccccc}
% \toprule
% $W$ & 128 & 256 & 512 & \cellcolor{gray!15}1024 & 2048 \\
% \midrule
% CIFAR-100-C & 23.36 & 26.68 & 28.62 & \cellcolor{gray!15}28.99 & 28.66 \\
% DomainNet   & 40.48 & 43.06 & 45.78 & \cellcolor{gray!15}46.64 & 45.34 \\
% \bottomrule
% \end{tabular}
% \end{table*}

\begin{figure*}[h!]
  \vspace{-\intextsep}
  \centering
  \includegraphics[width=0.90\linewidth]{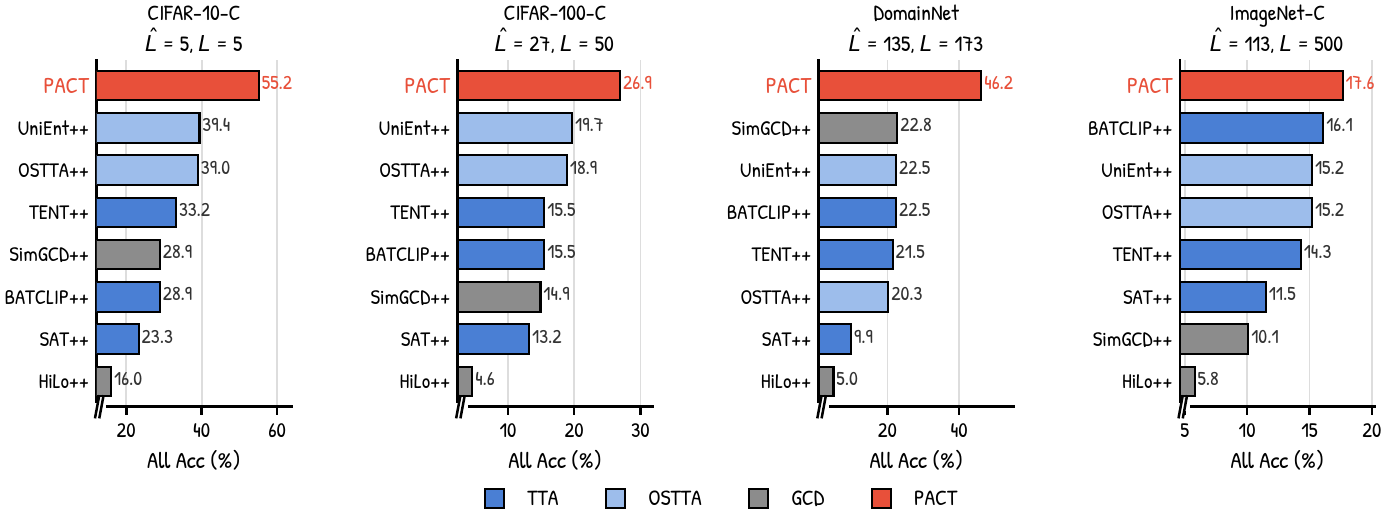}
  \caption{\textbf{TT-GCD under estimated $L$.} \texttt{All} accuracy (\%) when the novel budget is estimated once per stream and shared by every method (App.~\ref{app:estl}).}
  \label{fig:estl}
  \vspace{-1pt}
\end{figure*}

\mypar{Unknown number of novel classes.} 
\label{lest}
We follow the estimation procedure of GCD~\citep{vaze2022generalized} to estimate the number of novel classes. $k$-means is run over the zero-shot features for a range of cluster counts, the count maximizing the silhouette score is found by Brent's method, and the novel budget is $\hat{L}=\hat{k}-K$ with $K=|\Cknown|$ given by the setting. Fig.~\ref{fig:estl} reports the resulting accuracies. \ours\ remains first on the four benchmarks, \textbf{by $\boldsymbol{15.8}$ points on CIFAR-10-C, $\boldsymbol{7.3}$ on CIFAR-100-C and $\boldsymbol{23.4}$ on DomainNet}, and by $1.6$ on ImageNet-C.

% \mypar{Streaming stability.}
% Figure~\ref{fig:stream}(b) reports \texttt{All} accuracy across successive deciles of each DomainNet stream. Performance stabilizes after the initial portion and remains consistent throughout the stream, including the real and quickdraw domains with approximately $173$k samples each, demonstrating stable prototype adaptation over long streams.

% \mypar{Cost, Fig.~\ref{fig:stream}(b).} \textcolor{black}{The streaming stage of \ours\ involves no backward pass and thus it runs at $76$ ms per batch of 128 against $919$ to $1675$ ms for the baselines, i.e., \textbf{12-22$\times$ faster}, which back-propagate through the encoder on every batch. The one-off warmup takes $30$s of GPU time per stream (App.~\ref{app:cost})}

% \mypar{Long streams.} \textcolor{red}{Since the prototypes are updated on every batch, a wrong assignment could attract further samples and drift along the stream. Fig.~\ref{fig:stream}(c) shows that it does not. \texttt{All} accuracy is flat from the second decile on, including on real and quickdraw with about 173k samples each, and the second half of every domain scores $1.1$ to $2.8$ points above the first.}

\section{Conclusion}
We introduced TT-GCD, where a vision-language model must adapt to a shifted, unlabeled data stream where it must recognise known classes as well as discover the novel ones, without any source data or labels. Existing test-time adaptation and category discovery methods fail to address this problem. We introduced \ours\ which solves this with a one-shot re-alignment of the image-encoder LayerNorm parameters on a warmup buffer and a streaming assignment among visual prototypes.  \ours\ consistently outperforms seven adapted baselines across five benchmark datasets to effectively handle the task of TT-GCD.

% \section{AI USE STATEMENT}
% We used large language models (LLMs) to check the grammar and to help format tables and style figures. LLMs were not used to review the literature or design the method or to generate any mathematical proofs. The authors reviewed and verified all LLM-assisted content and take full responsibility for the paper.

\bibliographystyle{plainnat}
\bibliography{iclr2027_conference}

\newpage
\appendix
\appendix
\section{Baselines}
\label{app:baselines}

\subsection{Adaptation to TT-GCD}
\label{app:adapt}
TT-GCD asks for three things at once, (1) adaptation to distribution shift, (2) classification of known classes from text alone, and (3) discovery of novel categories, with no labeled data and no source dataset. No method operates under all three constraints. TTA methods assume a closed label set, open-set TTA detects and discards novel samples but does not categorize them, and every GCD method depends on labeled known-class data for supervised training of the classifier. Evaluating the setting therefore requires adapting existing methods to it.

We modify each baseline as its own assumptions allow. All seven adapted methods share the pipeline of Sec.~\ref{sec:setup}. The batch is scored against the known text anchors and partitioned by a two-component Gaussian mixture over the maximum cosine similarity, following the mixture formulation of UniEnt~\citep{gao2024unient} and extended here to the closed-set methods, which have no splitting mechanism of their own. The known-class objective of each method runs on the known subset, as the original objectives assume a closed label space, and the two open-set methods additionally apply their novel-entropy terms to the novel subset. Known samples are classified by argmax over the known text anchors, and novel samples are clustered by $k$-means with $k=L$ over the features produced at the end of the stream. The two GCD methods replace the last step by their own classifier over known and novel prototypes.

\begin{table*}[h]
\centering
\caption{\textbf{TT-GCD adaptation of each baseline, and \ours.} LN denotes LayerNorm affine parameters. The \colNovel{} column indicates how the test-time objective treats novel samples: \xmark\ not at all, $\uparrow$ entropy maximization only, \cmark\ modeled as categories with their own parameters.}
\label{tab:methods}
\fontsize{7.5}{9}\selectfont
\setlength{\tabcolsep}{5pt}
\begin{tabular}{@{}l l c l@{}}
\toprule
Method & Adapted parameters & \colNovel & Test-time objective \\
\midrule
\multicolumn{4}{@{}l}{\textbf{Test Time Adaptation}}\\
\midrule
TENT++~\citep{wangtent} & visual LN & \xmark
  & entropy minimization on known samples \\
BATCLIP++~\citep{maharana2025batclip} & visual + text LN & \xmark
  & entropy + image-to-text alignment + diversity \\
SAT++~\citep{mishra2026semanticanchortransportrobust} & visual LN & \xmark
  & Sinkhorn optimal-transport self-labeling \\
\addlinespace[2pt]
\midrule
\multicolumn{4}{@{}l}{\textbf{Open-Set Test Time Adaptation}}\\
\midrule
OSTTA++~\citep{lee2023ostta} & visual LN & $\uparrow$
  & known entropy min.\ $-$ novel entropy + diversity \\
UniEnt++~\citep{gao2024unient} & visual LN & $\uparrow$
  & OSTTA++ + clamped novel entropy maximization \\
\addlinespace[2pt]
\midrule
\multicolumn{4}{@{}l}{\textbf{Generalized Category Discovery}}\\
\midrule
SimGCD++~\citep{wen2023simgcd} & projection head + prototypes & \cmark
  & cross-view self-distillation \\
HiLo++~\citep{wang2025hilo} & prototype bank & \cmark
  & semantic entropy minimization \\
\midrule
\rowcolor{gray!12}
\ours\ (\textit{Ours}) & visual LN & \cmark
  & weighted cross-entropy + prototype assignment \\
\bottomrule
\end{tabular}
\end{table*}

\mypar{TTA.} TENT++, BATCLIP++ and SAT++ adapt only the LayerNorm affine parameters of the frozen encoder. TENT++ minimizes the Shannon entropy of the known-anchor logits. BATCLIP++ adapts the visual and text LayerNorms and adds an image-to-text alignment loss and an inter-class diversity loss. SAT++ performs Sinkhorn optimal-transport assignment on the known samples, distilled across the prompt templates.

\mypar{Open-Set TTA.} OSTTA++ adds a frozen reference model for confidence filtering, an entropy-maximization term on the novel subset and a batch-diversity term. UniEnt++ extends it with a clamped novel-entropy-maximization branch. Neither assigns novel samples to categories.

\mypar{GCD.} Both methods require labeled known-class data in their original form, which TT-GCD does not provide, so the known subset of the splitter with zero-shot pseudo-labels replaces the labeled set. SimGCD++ keeps the backbone frozen and learns a student projection head and novel prototypes by cross-view self-distillation, with the novel prototypes initialized from $k$-means over the early-stream buffer rather than at random. HiLo++ freezes the backbone and learns a joint prototype bank, known rows initialized from the text anchors and novel rows from $k$-means, by semantic entropy minimization. HiLo++ is the weakest baseline on CIFAR-100-C ($5.69$) and DomainNet ($5.37$). HiLo obtains its robustness to domain shift by training the backbone on cross-domain labeled data with a semantic-domain disentanglement objective, and TT-GCD provides neither a source dataset nor a training phase.

\begin{table*}[h!]
\centering
\caption{\textbf{Baseline hyperparameter sweeps} on the four held-out CIFAR-100-C corruptions. \emph{baseline} denotes the value of the original implementation. Best \texttt{All}(\%) is on the held-out corruptions and is not comparable to Table~\ref{tab:main}, which reports the 15 evaluation corruptions.}
\label{tab:bsweep}
\footnotesize
\setlength{\tabcolsep}{4pt}
\renewcommand{\arraystretch}{1.15}
\begin{tabular}{@{}l c p{7.6cm} c@{}}
\toprule
Method & \# & Swept values & Best All \\
\midrule
TENT++ & 1
  & LR $\{$baseline, $10^{-3.5}$, $10^{-2.5}\}$
  & 24.86 \\
\midrule
BATCLIP++ & 2
  & LR $\{$baseline, $10^{-3.5}$, $10^{-2.5}\}$; weight decay $\{$baseline, $0.01\}$
  & 17.86 \\
\midrule
SAT++ & 5
  & LR $\{$baseline, $10^{-4.5}$, $10^{-3.5}$, $10^{-2.5}\}$; temperature $\{$baseline,
    $0.005$, $0.05\}$; $\varepsilon$ $\{$baseline, $0.4$, $1.0\}$; Sinkhorn iterations
    $\{$baseline, $1$, $5\}$; multi-template $\{$yes, no$\}$
  & 26.09 \\
\midrule
OSTTA++ & 4
  & LR $\{$baseline, $10^{-3.5}$, $10^{-2.5}\}$; $\alpha_{\mathrm{div}}$ $\{$baseline, $0.1$,
    $0.5\}$; $\alpha_{\mathrm{ood}}$ $\{$baseline, $0.05$, $0.2\}$; steps $\{$baseline, $2\}$
  & 27.25 \\
\midrule
UniEnt++ & 4
  & LR $\{$baseline, $10^{-3.5}$, $10^{-2.5}\}$; $\alpha_{\mathrm{div}}$ $\{$baseline,
    $0.05$, $0.2\}$; $\alpha_{\mathrm{ood}}$ $\{$baseline, $0.025$, $0.1\}$; steps
    $\{$baseline, $2\}$
  & 22.75 \\
\midrule
SimGCD++ & 5
  & LR $\{$baseline, $10^{-2.5}$, $10^{-1.5}\}$; EMA $\{$baseline, $0.99\}$;
    $w_{\mathrm{cons}}$ $\{$baseline, $0.25$, $1.0\}$; $w_{\mathrm{contr}}$ $\{$baseline,
    $0.75$, $3.0\}$; $w_{\mathrm{div}}$ $\{$baseline, $0.005$, $0.05\}$
  & 19.68 \\
\midrule
HiLo++ & 3
  & LR $\{$baseline, $10^{-2.5}$, $10^{-1.5}\}$; temperature $\{$baseline, $0.05$, $0.2\}$;
    $w_{\mathrm{sem}}$ $\{$baseline, $0.25$, $1.0\}$
  & 7.07 \\
\bottomrule
\end{tabular}
\end{table*}

\subsection{Hyperparameter selection for the baselines}
\label{app:bsweep}
Every baseline receives a sweep on the four held-out CIFAR-100-C corruptions, covering its learning rate and the method-specific coefficients that its authors expose. Values are varied one at a time around the original implementation, and the selected configuration is then fixed for every benchmark. Table~\ref{tab:bsweep} reports the grids. The number of swept hyperparameters ranges from one to five across the baselines. Accuracies in the last column are measured on the held-out corruptions and are not comparable to Table~\ref{tab:main}, which uses a different, non-overlapping set of 15 corruptions.

\section{Datasets and protocol}
\label{app:data}
\mypar{Corruption benchmarks.} CIFAR-10-C, CIFAR-100-C and ImageNet-C apply 15 corruption types to the test set of the corresponding clean dataset:
\begin{wraptable}{r}{0.40\textwidth}
\centering
\caption{\textbf{Benchmarks.} Each corruption type forms one test stream.}
\label{tab:splits}
\footnotesize
\setlength{\tabcolsep}{1pt}
\renewcommand{\arraystretch}{1.0}
\begin{tabular}{@{}l c ccc@{}}
\toprule
Dataset & Streams & $|\mathcal{C}|$ & $K$ & $L$ \\
\midrule
\multicolumn{5}{@{}l}{\emph{Coarse-grained corruptions}}\\
\midrule
CIFAR-10-C  & 15 & 10   & 5   & 5   \\
CIFAR-100-C & 15 & 100  & 50  & 50  \\
ImageNet-C  & 15 & 1000 & 500 & 500 \\
\midrule
\multicolumn{5}{@{}l}{\emph{Fine-grained corruptions}}\\
\midrule
CUB-C       & 7  & 200  & 100 & 100 \\
\midrule
\multicolumn{5}{@{}l}{\emph{Natural shift}}\\
\midrule
DomainNet   & 6  & 345  & 172 & 173 \\
\bottomrule
\end{tabular}
\vspace{-1.5\intextsep}
\end{wraptable}
\texttt{gaussian\_noise}, \texttt{shot\_noise}, \texttt{impulse\_noise}, \texttt{defocus\_blur}, \texttt{glass\_blur}, \texttt{motion\_blur}, \texttt{zoom\_blur}, \texttt{snow}, \texttt{frost}, \texttt{fog}, \texttt{brightness}, \texttt{contrast}, \texttt{elastic\_transform}, \texttt{pixelate} and \texttt{jpeg\_compression}. We use severity 5 throughout and treat each corruption as a separate test set. The four extra corruptions of CIFAR-100-C (\texttt{gaussian\_blur}, \texttt{saturate}, \texttt{spatter}, \texttt{speckle\_noise}) are used only for hyperparameter selection. CUB-C follows the corruption protocol of HiLo~\citep{wang2025hilo} on CUB~\citep{WahCUB_200_2011} with seven corruptions at severity 5: \texttt{fog}, \texttt{frost}, \texttt{gaussian\_noise}, \texttt{impulse\_noise}, \texttt{shot\_noise}, \texttt{snow} and \texttt{zoom\_blur}.

\mypar{DomainNet.} DomainNet~\citep{peng2019momentmatchingmultisourcedomain} contains 345 classes in six domains (clipart, infograph, painting, quickdraw, real, sketch). Each domain is one test stream and no domain is used as a source.

\mypar{Class splits.} For every dataset the class set is partitioned into $\Cknown$ and $\Cnovel$ by a seeded random permutation, shared by all methods and all corruptions or domains of that dataset. Table~\ref{tab:splits} lists the resulting class counts and number of streams.

\section{Implementation details of \ours}
\label{app:impl}

\subsection{Configuration}
\label{app:config}

\begin{table}[h]
\centering
\caption{\textbf{\ours\ configuration,} same for every benchmark.}
\label{tab:config}
\fontsize{7.5}{9}\selectfont
\setlength{\tabcolsep}{5pt}
\begin{tabular}{@{}l l c@{}}
\toprule
Parameter & Meaning & Value\\
\midrule
\multicolumn{3}{@{}l}{\textbf{Re-alignment}}\\
\midrule
$W$ & buffer size & 1024\\
$E$ & epochs & 30\\
LR  & learning rate & $10^{-2}$ \\
\addlinespace[2pt]
\midrule
\multicolumn{3}{@{}l}{\textbf{Streaming}}\\
\midrule
$e_{\min}$ & known activation threshold & 1.5\\
$\eta$ & known prototype rate & 0.05\\
$\lambda$ & novel prototype rate & 0.10\\
$|\mathcal{B}_t|$ & batch size & 128\\
\bottomrule
\end{tabular}
\end{table}

Table~\ref{tab:config} lists the configuration used on every benchmark. Re-alignment is performed once per stream on $W$ samples for $E$ epochs, adapting the LayerNorm affine parameters of the image encoder, after which they are frozen. The text encoder is never updated, so the $K$ anchors are encoded once per stream. The streaming stage adds one encoder forward pass and a $(K{+}L)$-way dot product per sample, so its cost over the stream is that of a single inference pass with a linear head.

\mypar{Warmup learning rate.} The warmup learning rate is selected by a sweep on the four extra CIFAR-100-C corruptions of App.~\ref{app:data}, where All is $34.60$, $34.27$, $35.09$ and $33.54$ for $10^{-3}$, $3{\times}10^{-3}$, $10^{-2}$ and $3{\times}10^{-2}$, respectively. The four values lie within $1.6$ points of each other, and above $10^{-2}$ the warmup over-adapts. The remaining values of Table~\ref{tab:config} are shared across all benchmarks, and App.~\ref{app:sens} reports their sensitivity.

\subsection{Computational cost}
\label{app:cost}
Table~\ref{tab:cost} reports the steady-state time per batch of 128 and the peak GPU memory of every method on one H100 3g.40gb slice, for one CIFAR-100-C corruption and for DomainNet clipart. The streaming stage of \ours\ involves no backward pass, so it costs $60$ ms per batch on CIFAR-100-C and $76$ ms on DomainNet at $1.1$ GB, against $919$ to $1771$ ms for the baselines, which back-propagate through the encoder on every batch. The warmup is a one-off cost of $30$ s of GPU time per stream, whose wall-clock time ($38$ s and $60$ s) includes re-decoding the buffer at every epoch. Warmup included, a CIFAR-100-C stream costs $35$ s of GPU time and DomainNet clipart $59$ s, against $73$ s and $345$ s for HiLo++, the fastest baseline.

\begin{table}[h]
\centering
\caption{\textbf{Cost per batch.} Steady-state ms per batch of 128 and peak GPU memory (GB) on one H100 3g.40gb slice.}
\label{tab:cost}
\fontsize{7.5}{9}\selectfont
\setlength{\tabcolsep}{5pt}
\begin{tabular}{@{}l cc cc@{}}
\toprule
 & \multicolumn{2}{c}{\textbf{CIFAR-100-C}} & \multicolumn{2}{c}{\textbf{DomainNet}} \\
\cmidrule(lr){2-3}\cmidrule(lr){4-5}
Method & ms/batch & peak GB & ms/batch & peak GB \\
\midrule
TENT++    & 1106 & 8.7  & 1050 & 10.3 \\
BATCLIP++ & 1066 & 8.4  & 1181 & 10.5 \\
SAT++     & 1014 & 5.6  & 1085 & 10.7 \\
OSTTA++   & 1771 & 18.0 & 1673 & 18.1 \\
UniEnt++  & 1758 & 18.0 & 1675 & 18.1 \\
SimGCD++  & 1558 & 2.5  & 1540 & 2.6  \\
HiLo++    & 920  & 3.3  & 919  & 3.4  \\
\midrule
\rowcolor{gray!12}
\ours, stream (forward $+$ prototype ops) & 60 {\scriptsize(46 $+$ 14)} & 1.1 & 76 {\scriptsize(46 $+$ 30)} & 1.1 \\
\rowcolor{gray!12}
\ours, warmup (one-off per stream) & 30\,s GPU {\scriptsize(38\,s wall)} & 8.5 & 30\,s GPU {\scriptsize(60\,s wall)} & 8.5 \\
\bottomrule
\end{tabular}
\end{table}

\section{Additional results}
\label{app:results}

\subsection{DomainNet: \colKnown{} and \colNovel{} per domain}
\label{app:dn}
Tables~\ref{tab:dn_known} and~\ref{tab:dn_novel} give the label-space split behind the All accuracies of Table~\ref{tab:domainnet}. \ours\ is the best method on both halves of the label space in all six domains. Its margin over the strongest baseline ranges from $10.6$ (Infograph) to $17.0$ (Clipart) points on \colKnown{}, and from $12.0$ (Infograph) to $33.5$ (Real) on \colNovel{}, so the gain on the novel half exceeds the gain on the known half in every domain. 

\begin{table*}[h!]
\centering
\caption{DomainNet, \colKnown{} accuracy (\%) per domain, mean over 3 seeds with the standard deviation. Best in \textbf{bold}, second best \underline{underlined}.}
\label{tab:dn_known}
\footnotesize
\setlength{\tabcolsep}{4pt}
\begin{tabular}{@{}l cccccc@{}}
\toprule
Method & Clipart & Infograph & Painting & Quickdraw & Real & Sketch \\
\midrule
TENT++ & \vc{\underline{49.29}}{.43} & \vc{29.16}{.21} & \vc{\underline{46.94}}{.06} & \vc{0.82}{.30} & \vc{\underline{57.02}}{.14} & \vc{41.77}{.09} \\
BATCLIP++ & \vc{37.94}{.21} & \vc{21.84}{.07} & \vc{36.56}{.22} & \vc{7.29}{.11} & \vc{37.75}{1.06} & \vc{33.21}{.28} \\
SAT++ & \vc{16.79}{.14} & \vc{13.51}{.52} & \vc{11.53}{1.37} & \vc{2.57}{.28} & \vc{9.32}{.40} & \vc{13.59}{.89} \\
OSTTA++ & \vc{46.31}{.54} & \vc{\underline{31.50}}{.74} & \vc{43.63}{.93} & \vc{\underline{8.68}}{.35} & \vc{45.09}{.86} & \vc{40.93}{.15} \\
UniEnt++ & \vc{48.79}{.19} & \vc{31.41}{.54} & \vc{45.98}{.36} & \vc{7.98}{.40} & \vc{51.87}{.91} & \vc{\underline{42.35}}{.86} \\
SimGCD++ & \vc{23.70}{23.75} & \vc{10.65}{4.88} & \vc{22.43}{27.54} & \vc{7.30}{.54} & \vc{25.42}{23.25} & \vc{10.95}{3.69} \\
HiLo++ & \vc{7.34}{.53} & \vc{5.17}{1.03} & \vc{6.35}{.10} & \vc{3.16}{.05} & \vc{7.06}{.60} & \vc{5.08}{.45} \\
\midrule
\rowcolor{gray!12}
\ours\ (\textit{Ours}) & \vc{\textbf{66.29}}{1.44} & \vc{\textbf{42.12}}{.16} & \vc{\textbf{62.37}}{1.96} & \vc{\textbf{20.96}}{.60} & \vc{\textbf{73.01}}{1.06} & \vc{\textbf{57.12}}{1.21} \\
\bottomrule
\end{tabular}
\end{table*}

\begin{table*}[h!]
\centering
\caption{DomainNet, \colNovel{} accuracy (\%) per domain, mean over 3 seeds with the standard deviation. Best in \textbf{bold}, second best \underline{underlined}.}
\label{tab:dn_novel}
\footnotesize
\setlength{\tabcolsep}{4pt}
\begin{tabular}{@{}l cccccc@{}}
\toprule
Method & Clipart & Infograph & Painting & Quickdraw & Real & Sketch \\
\midrule
TENT++ & \vc{6.65}{.07} & \vc{5.93}{.13} & \vc{7.18}{.11} & \vc{0.23}{.30} & \vc{6.22}{.08} & \vc{7.22}{.29} \\
BATCLIP++ & \vc{\underline{19.95}}{.12} & \vc{11.55}{.22} & \vc{\underline{19.74}}{.41} & \vc{\underline{4.80}}{.13} & \vc{\underline{23.48}}{.93} & \vc{\underline{17.34}}{.53} \\
SAT++ & \vc{11.46}{.57} & \vc{\underline{15.04}}{.36} & \vc{8.11}{.39} & \vc{1.84}{.29} & \vc{6.53}{.50} & \vc{8.49}{.63} \\
OSTTA++ & \vc{6.68}{.18} & \vc{7.13}{.61} & \vc{7.95}{.17} & \vc{2.77}{.19} & \vc{9.25}{.49} & \vc{8.62}{.11} \\
UniEnt++ & \vc{6.53}{.06} & \vc{8.24}{.76} & \vc{7.86}{.68} & \vc{3.05}{.43} & \vc{8.50}{.47} & \vc{9.01}{1.13} \\
SimGCD++ & \vc{10.28}{7.31} & \vc{10.42}{4.26} & \vc{8.35}{5.43} & \vc{3.17}{.51} & \vc{10.49}{6.55} & \vc{6.15}{1.02} \\
HiLo++ & \vc{4.89}{1.19} & \vc{6.09}{.78} & \vc{7.16}{.28} & \vc{2.17}{.52} & \vc{4.81}{.63} & \vc{5.14}{.37} \\
\midrule
\rowcolor{gray!12}
\ours\ (\textit{Ours}) & \vc{\textbf{47.80}}{2.51} & \vc{\textbf{27.04}}{1.44} & \vc{\textbf{46.49}}{1.61} & \vc{\textbf{20.53}}{.58} & \vc{\textbf{57.01}}{.74} & \vc{\textbf{42.44}}{.60} \\
\bottomrule
\end{tabular}
\end{table*}

\subsection{CUB-C: \colKnown{} and \colNovel{} per backbone}
\label{app:cub}
Table~\ref{tab:cub_full} expands Table~\ref{tab:cub} with the label-space split. \ours\ obtains the best \colNovel{} with CLIP L/14 and BioCLIP-2, and the best \colKnown{} with both CLIP backbones. The two exceptions are SimGCD++ on \colNovel{} with CLIP B/16 ($15.75$ against $15.09$) and TENT++ on \colKnown{} with BioCLIP-2 ($59.26$ against $56.66$). A stronger backbone improves known-class recognition for the TTA baselines without helping them discover novel categories. \ours\ improves on both halves, by $29.9$ points on \colKnown{} and $13.9$ on \colNovel{}.

\begin{table}[h]
\centering
\caption{Backbone study on CUB-C (7 corruption types). Clustering accuracy (\%). All is in Table~\ref{tab:cub}.}
\label{tab:cub_full}
\footnotesize
\setlength{\tabcolsep}{5pt}
\begin{tabular}{@{}l ccc ccc ccc@{}}
\toprule
& \multicolumn{3}{c}{\textbf{CLIP B/16}} & \multicolumn{3}{c}{\textbf{CLIP L/14}}
& \multicolumn{3}{c}{\textbf{BioCLIP-2 L/14}} \\
\cmidrule(lr){2-4}\cmidrule(lr){5-7}\cmidrule(lr){8-10}
Method & All & \colKnown & \colNovel & All & \colKnown & \colNovel
       & All & \colKnown & \colNovel \\
\midrule
TENT++    & 9.54 & \underline{12.03} & 7.04 & 15.01 & \underline{21.81} & 8.21
          & \underline{31.97} & \textbf{59.26} & 4.73 \\
BATCLIP++ & 11.57 & 11.37 & 11.77 & 15.52 & 17.03 & 14.01 & 28.63 & 45.70 & 11.60 \\
SAT++     & 9.11 & 9.05 & 9.18 & 10.42 & 11.37 & 9.48 & 26.22 & 40.08 & 12.40 \\
OSTTA++   & 8.86 & 11.58 & 6.15 & 14.00 & 20.92 & 7.11 & 28.82 & \underline{53.15} & 4.53 \\
UniEnt++  & 7.57 & 9.72 & 5.42 & 13.96 & 18.60 & 9.33 & 20.58 & 34.33 & 6.85 \\
SimGCD++  & \underline{13.74} & 11.72 & \underline{15.75} & \underline{17.85} & 16.42
          & \underline{19.29} & 24.76 & 30.66 & \underline{18.87} \\
HiLo++    & 5.83 & 5.77 & 5.90 & 7.08 & 6.84 & 7.32 & 9.57 & 11.04 & 8.10 \\
\midrule
\rowcolor{gray!12}
\ours\ (\textit{Ours}) & \textbf{18.20} & \textbf{21.32} & \underline{15.09}
          & \textbf{23.59} & \textbf{26.74} & \textbf{20.44}
          & \textbf{45.48} & \underline{56.66} & \textbf{34.32} \\
\bottomrule
\end{tabular}
\end{table}

\subsection{Estimated $L$: \colKnown{} and \colNovel}
\label{app:estl}

Table~\ref{tab:estl_full} gives the accuracies behind Fig.~\ref{fig:estl}. 

\begin{table*}[h!]
\centering
\caption{\textbf{TT-GCD under estimated $L$.} Clustering accuracy (\%). Best in \textbf{bold}, second best \underline{underlined}.}
\label{tab:estl_full}
\fontsize{7.5}{9}\selectfont
\setlength{\tabcolsep}{2pt}
\begin{tabular}{@{}l ccc ccc ccc ccc@{}}
\toprule
& \multicolumn{3}{c}{\textbf{CIFAR-10-C}} & \multicolumn{3}{c}{\textbf{CIFAR-100-C}}
& \multicolumn{3}{c}{\textbf{DomainNet}} & \multicolumn{3}{c}{\textbf{ImageNet-C}} \\
& \multicolumn{3}{c}{$\hat{L}{=}5$, $L{=}5$} & \multicolumn{3}{c}{$\hat{L}{=}27$, $L{=}50$}
& \multicolumn{3}{c}{$\hat{L}{=}135$, $L{=}173$} & \multicolumn{3}{c}{$\hat{L}{=}113$, $L{=}500$} \\
\cmidrule(lr){2-4}\cmidrule(lr){5-7}\cmidrule(lr){8-10}\cmidrule(lr){11-13}
Method & All & \colKnown & \colNovel & All & \colKnown & \colNovel
       & All & \colKnown & \colNovel & All & \colKnown & \colNovel \\
\midrule
\multicolumn{13}{@{}l}{\textbf{Test Time Adaptation}}\\
\midrule
TENT++    & 33.20 & 59.89 & 6.50 & 15.47 & 17.02 & 13.92 & 21.50 & 37.48 & 5.77 & 14.25 & 20.29 & 8.21 \\
BATCLIP++ & 28.90 & 46.73 & 11.08 & 15.54 & 15.34 & 15.74 & 22.53 & 28.98 & \underline{16.19} & \underline{16.07} & 18.83 & \underline{13.30} \\
SAT++     & 23.27 & 34.20 & 12.33 & 13.19 & 13.71 & 12.67 & 9.85 & 11.15 & 8.57 & 11.48 & 13.17 & 9.80 \\
\addlinespace[2pt]
\midrule
\multicolumn{13}{@{}l}{\textbf{Open-Set Test Time Adaptation}}\\
\midrule
OSTTA++   & 39.05 & \underline{70.16} & 7.94 & 18.95 & 20.64 & 17.25 & 20.27 & 33.66 & 7.11 & 15.19 & \underline{21.93} & 8.45 \\
UniEnt++  & \underline{39.40} & \textbf{71.54} & 7.26 & \underline{19.66} & \underline{21.25} & \underline{18.07} & 22.49 & \underline{38.17} & 7.02 & 15.16 & \textbf{22.13} & 8.20 \\
\addlinespace[2pt]
\midrule
\multicolumn{13}{@{}l}{\textbf{Generalized Category Discovery}}\\
\midrule
SimGCD++  & 28.89 & 48.17 & 9.62 & 14.87 & 14.89 & 14.84 & \underline{22.80} & 33.56 & 12.19 & 10.05 & 10.95 & 9.15 \\
HiLo++    & 15.97 & 16.81 & \underline{15.12} & 4.58 & 3.54 & 5.61 & 5.04 & 5.75 & 4.33 & 5.77 & 6.19 & 5.34 \\
\midrule
\rowcolor{gray!12}
\ours\ (\textit{Ours}) & \textbf{55.18} & 66.96 & \textbf{43.41} & \textbf{26.95} & \textbf{34.33} & \textbf{19.57}
          & \textbf{46.21} & \textbf{53.55} & \textbf{38.96} & \textbf{17.64} & 21.22 & \textbf{14.07} \\
\bottomrule
\end{tabular}
\end{table*}

\subsection{Sensitivity analyses}
\label{app:sens}
The sweeps below vary one hyperparameter at a time around the configuration of Table~\ref{tab:config}, on the evaluation corruptions and domains. They are run after the configuration was fixed and no value reported in the main text is selected from them.

\mypar{Prototype rates.} Table~\ref{tab:rates} varies $\eta$ and $\lambda$ around the value used for our runs. The two values nearest to it, $\eta=0.10$ and $\lambda=0.20$, change All by $+0.08$ and $+0.11$ points on CIFAR-100-C and by $-0.25$ and $-0.42$ on DomainNet, within the standard deviation reported in Tables~\ref{tab:main} and~\ref{tab:domainnet}. Over the full range of $\lambda$, All moves by at most $1.2$ and $2.4$ points. Raising either rate above its fixed value raises \colKnown{} and lowers \colNovel{}. $\eta=0.60$ takes DomainNet \colKnown{} from $53.70$ to $57.27$ and \colNovel{} from $39.68$ to $23.40$. 

\begin{table*}[h]
\centering
\caption{\textbf{Prototype update rates,} varied one at a time around the fixed value. All, \colKnown{} and \colNovel{} (\%), mean over CIFAR-100-C corruptions and DomainNet domains.}
\label{tab:rates}
\fontsize{7.5}{9}\selectfont
\setlength{\tabcolsep}{5pt}
\begin{tabular}{@{}l ccc ccc@{}}
\toprule
& \multicolumn{3}{c}{\textbf{CIFAR-100-C}} & \multicolumn{3}{c}{\textbf{DomainNet}} \\
\cmidrule(lr){2-4}\cmidrule(lr){5-7}
Rate & All & \colKnown & \colNovel & All & \colKnown & \colNovel \\
\midrule
\multicolumn{7}{@{}l}{\textbf{Known rate} $\eta$}\\
\midrule
\rowcolor{gray!12}
0.05 & 28.99 & 33.74 & 24.25 & 46.64 & 53.70 & 39.68 \\
0.10 & 29.07 & 34.62 & 23.51 & 46.39 & 54.87 & 38.01 \\
0.30 & 28.23 & 35.47 & 20.99 & 43.65 & 56.74 & 30.70 \\
0.60 & 27.22 & 36.09 & 18.35 & 40.26 & 57.27 & 23.40 \\
\addlinespace[2pt]
\midrule
\multicolumn{7}{@{}l}{\textbf{Novel rate} $\lambda$}\\
\midrule
0.05 & 28.60 & 33.65 & 23.54 & 46.16 & 53.45 & 38.96 \\
\rowcolor{gray!12}
0.10 & 28.99 & 33.74 & 24.25 & 46.64 & 53.70 & 39.68 \\
0.20 & 29.10 & 34.35 & 23.85 & 46.22 & 54.34 & 38.18 \\
0.40 & 27.94 & 35.29 & 20.59 & 44.25 & 55.22 & 33.37 \\
\bottomrule
\end{tabular}
\end{table*}

\begin{figure*}[h]
  \centering
  \includegraphics[width=0.80\linewidth]{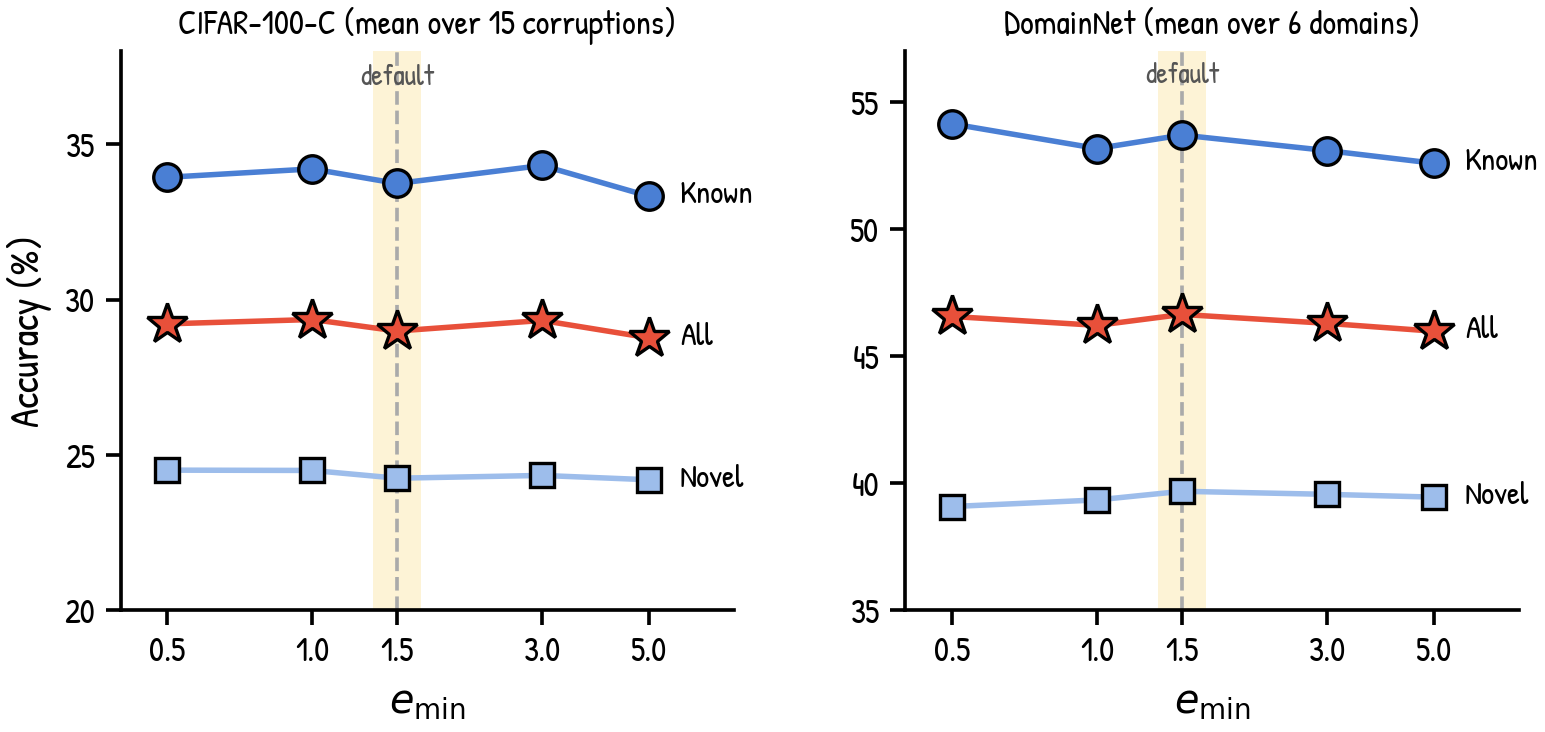}
  \caption{Activation threshold $e_{\min}$ on CIFAR-100-C and DomainNet.}
  \label{fig:emin}
\end{figure*}

\mypar{Activation threshold.} Fig.~\ref{fig:emin} varies $e_{\min}$ from $0.5$ to $5$. All, \colKnown{} and \colNovel{} move by at most $1.6$ points across the range on both benchmarks.

% \section{One-shot vs.\ continual re-alignment}
% \label{app:continual}
% \begin{wrapfigure}[14]{r}{0.40\textwidth}
% \vspace{-1.2\intextsep}
% \centering
% \includegraphics[width=\linewidth]{OnGoingVersion/figures/fig_continual.pdf}
% \caption{\textbf{One-shot vs.\ continual re-alignment.} Dashed: \ours, encoder frozen after the warmup. $10^{-2}$ is the warmup learning rate.}
% \label{fig:continual}
% \vspace{-1.0\intextsep}
% \end{wrapfigure}
% \ours\ adapts the LayerNorm parameters once, on the warmup buffer, and keeps them fixed during streaming. Fig.~\ref{fig:continual} assesses whether continuing the adaptation on every streaming batch, with the same objective, brings further gains. Each batch is predicted before the model is updated on it. Continual adaptation at best matches the one-shot schedule where it stays within $0.2$ points of it at a streaming learning rate of $10^{-5}$, loses $10$ points on DomainNet at $10^{-4}$, and collapses to $5.3$ on CIFAR-100-C and $2.0$ on DomainNet at $10^{-2}$, the warmup learning rate. We attribute this to the pseudo-labels. During warmup they are computed once with the initial model, whereas during streaming they come from the model being adapted, so its errors reinforce themselves (Sec.~\ref{sec:align}).

\end{document}